\pdfoutput=1

\documentclass[11pt]{article}

\usepackage{acl}

\usepackage{algorithmicx}
\usepackage{amssymb}
\usepackage{adjustbox}            

\usepackage{times}
\usepackage{latexsym}
\usepackage{dirtytalk}  
\usepackage{listings}

\usepackage{amsmath}
\usepackage{array,multirow,graphicx}
\usepackage{float}
\usepackage{subcaption}

\usepackage[T1]{fontenc}
\usepackage[utf8]{inputenc}
\usepackage{microtype}
\usepackage{booktabs}
\usepackage{fancyvrb}
\usepackage{caption}
\usepackage{csquotes}
\usepackage{appendix}
\usepackage{titlesec}
\usepackage[section]{placeins}

\usepackage{algorithm}
\usepackage[noend]{algpseudocode}

\usepackage{stfloats}

\usepackage{hyperref}
\usepackage{tcolorbox}
\tcbuselibrary{listingsutf8, breakable}

\usepackage{graphicx}
\newcommand{\absi}{\raisebox{-0.2pt}{\includegraphics[scale=0.5]{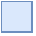}}}
\newcommand{\abso}{\raisebox{-0.2pt}{\includegraphics[scale=0.5]{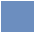}}}
\newcommand{\ver}{\raisebox{-0.2pt}{\includegraphics[scale=0.5]{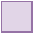}}}

\newcommand{\annptators}{\raisebox{-0.2pt}{\includegraphics[scale=0.5]{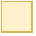}}}
\newcommand{\prompt}{\raisebox{-0.2pt}{\includegraphics[scale=0.5]{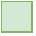}}}
\newcommand{\model}{\raisebox{-0.2pt}{\includegraphics[scale=0.5]{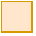}}}

\newcommand{\si}{\raisebox{-0.2pt}{\includegraphics[scale=0.5]{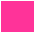}}}
\newcommand{\li}{\raisebox{-0.2pt}{\includegraphics[scale=0.5]{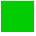}}}

\newcommand{\eone}{\raisebox{-0.2pt}{\includegraphics[scale=0.5]{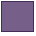}}}
\newcommand{\rone}{\raisebox{-0.1pt}{\includegraphics[scale=0.6]{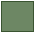}}}
\newcommand{\etwo}{\raisebox{-0.2pt}{\includegraphics[scale=0.5]{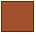}}}

\title{Automated Knowledge Graph Construction using Large Language Models and Sentence Complexity Modelling}

\author{
Sydney Anuyah\textsuperscript{\dag} \quad Mehedi Mahmud Kaushik\textsuperscript{\dag} \quad Krishna Dwarampudi\textsuperscript{\ddag} \\ 
\smallskip \bf Rakesh Shiradkar\textsuperscript{\S} \quad Arjan Durresi\textsuperscript{\dag} \quad Sunandan Chakraborty \textsuperscript{\dag}\\
\begin{tabular}{ccc}
      \textsuperscript{\dag}  Indiana University & \textsuperscript{\ddag}Purdue University & \textsuperscript{\S}Department of Biomedical Engineering\\
     Indianapolis, IN  &  Indianapolis, IN & and Informatics, Indiana University, Indianapolis, IN\\
\end{tabular} \\
\texttt{\small\{sanuyah, mekaush, srdwar, rshirad, adurresi, sunchak \}@iu.edu}}

\begin{document}

\maketitle
\renewcommand{\thefootnote}{\fnsymbol{footnote}}

\begin{abstract}
We introduce CoDe-KG, an open-source, end-to-end pipeline for extracting sentence-level knowledge graphs by combining robust coreference resolution with syntactic sentence decomposition. Using our model, we contribute a dataset of over 150 000 knowledge triples, which is open source. We also contribute a training corpus of 7248 rows for sentence complexity, 190 rows of gold human annotations for co-reference resolution using open source lung-cancer abstracts from PubMed, 900 rows of gold human annotations for sentence conversion policies, and 398 triples of gold human annotations. We systematically select optimal prompt-model pairs across five complexity categories, showing that hybrid chain-of-thought and few-shot prompting yields up to 99.8\% exact-match accuracy on sentence simplification. On relation extraction (RE), our pipeline achieves 65.8\% macro-F1 on REBEL, an 8-point gain over the prior state of the art, and 75.7\% micro-F1 on WebNLG2, while matching or exceeding performance on Wiki-NRE and CaRB. Ablation studies demonstrate that integrating coreference and decomposition increases recall on rare relations by over 20\%. Code and dataset are available at \href{https://github.com/KaushikMahmud/CoDe-KG_EMNLP_2025}{https://github.com/KaushikMahmud/CoDe-KG\_EMNLP\_2025}.

\end{abstract}

\section{Introduction and Background}
\label{sec:intro}

One way to represent data is through knowledge graphs (KGs) \cite{hogan2021knowledge}. KGs have transformed the way data is organized and by leveraging complex network chains, we have been able to explore complex fields like causality in different domains \cite{friedman2022unstructured, maclean2021knowledge, naser2022causality}. 

With the recent advancements in Natural Language Processing (NLP) using Large Language Models (LLMs), KGs have become instrumental, both as knowledge bases and in finetuning these large models \cite{pan2024unifying}. One of such advantages is in the creation of domain-specific ontologies \cite{karim2023large, chandak2023building} closely associated with creating new reasoning and inference methods \cite{kau2024combining, zhang2024knowledge}.

Previous research has built the foundational concepts of KGs, which include the models used in creating these graphs and their representation \cite{hogan2021knowledge}. Automated KG constructions \cite{zhong2023comprehensive} and representation learning \cite{ji2021survey} have defined major stages in building a KG: from knowledge acquisition and semantic table interpretation \cite{liu2023tabular} to entity extraction--covering Named Entity Recognition (NER), Named Entity Disambiguation (NED), and Named Entity Linking (NEL) \cite{al2020named}. These studies and many more provide a system in which unstructured text can be transformed into an organized corpus of interlinked entities. Secondly, domain techniques such as graph knowledge distillation \cite{tian2023knowledge} and embedding schemes \cite{cao2024knowledge} have helped reinforce the ability to compress, optimize, and represent KGs which are then utilized in downstream applications. The goal of event KGs \cite{guan2022event} and explainable artificial intelligence (AI)  on KGs \cite{schramm2023comprehensible} is to empathically ensure that models not only have to be efficient but also interpretable \cite{kaur2022trustworthy}.

The challenges we are tackling is two fold: (1) We have a large volume of unstructured text data, and because it varies in structure, writing style, and vocabulary across different domains, it has become harder to parse, and (2) Many automated pipelines for KG creation are not really automated, as some are heavily prompt reliant on the end user \cite{buehler2024accelerating}, others have issues of handling noisy datasets \cite{zhang2023saka}. This is why we are researching a one-size-fits-all open-source model  framework that could help in knowledge extraction irrespective of our data. Through the typical structure of the English Language, it is possible to extract relationships through verb usage and clauses. This leads us to the first research question. \textbf{RQ1:} Can sentence modelling be used to effectively create KGs that rival other methods? We also compare our method to the popular closed-source AI model: GPT 4 series which is renowned for parsing academic literature, and we design evaluation prompts to benchmark their performance against ours. This can be summarized as \textbf{RQ2:} can an open‐source, LLM model using the sentence semantics approach reliably construct KGs from raw texts?
Our contributions are as follows:
\begin{itemize}
    \item We introduce a novel sentence-semantic framework  for relation extraction (RE) and KG construction, borrowing from linguistic theory and semantic parsing. This idea though common, to the best of our knowledge has been under-explored in mainstream NLP information extraction pipelines. The novelty of our work lies in the integration of multiple frameworks rather than just one task. Our method explicitly models semantic sentence types (e.g., complex (CX), compound (CD), and compound-complex (CC) forms) as the foundation for extracting  knowledge triples. Each triples is a simple, three-part structure $(\text{entity}_1,\,\text{relationship},\,\text{entity}_2)$ used to represent a single fact in a KG. 
    \item We also explore diverse prompting strategies across our pipeline, including Chain-of-Thought (CoT) reasoning, Few-Shot In-Context Learning (FICL), and Zero-Shot General Instruction prompting (GIP), and empirically demonstrate their varying contributions to structural decomposition. To support this architecture, we release a suite of open-source resources:
    \begin{enumerate}
        \item A 7248-row dataset that categorizes and maps diverse sentence semantics aligned to our model’s decomposition strategy (complex, compound, compound-complex, simple and incomplete sentence).
        \item A gold-standard co-reference resolution corpus comprising 190 PubMed lung-cancer abstracts annotated by four domain experts.
        \item A 900-sample sentence transformation dataset, consisting of 300 annotated examples each for converting complex, compound, and compound-complex sentences into simple, extractable forms.
        \item A machine-generated KG corpus of over 150,000 structured triples, created using our full end-to-end pipeline.
    \end{enumerate}
\end{itemize}

\section{Background}
\subsection{Sentence Semantic Modelling for Knowledge Extraction}
\label{sec:sentence_modeling}

\begin{table*}[ht]
  \centering
  \small
  \begin{tabular}{lccccc}
    \toprule
    Method & Sentence Decomp.\ & Coref Res.\ & Open-Source & Domain-Agnostic & Eval Scripts \\
    \midrule
    GraphRAG  \cite{han2024retrieval}    & $\times$        & $\times$      & $\times$       & \checkmark     & $\times$       \\
    EDC  \cite{zhang2024extract}         & $\times$        & $\times$      & \checkmark     & \checkmark     & \checkmark     \\
    GKG-LLM \cite{zhang2025gkg}      & $\times$        & \checkmark    & \checkmark     & \checkmark     & \checkmark     \\
    Neo4j LLM-KG  \cite{bharti2024enhancing} & $\times$        & $\times$      & $\times$    & \checkmark     & $\times$       \\
    KGGen (2025) \cite{mo2025kggen}        & $\times$        & $\times$      & \checkmark     & \checkmark     & \checkmark     \\
    Our Pipeline        & \checkmark      & \checkmark    & \checkmark     & \checkmark     & \checkmark     \\
    \bottomrule
  \end{tabular}
  \caption{Current LLM-induced KG Methods Comparison}
  \label{tab:comparison}
\end{table*}

Sentence semantic modelling involves organizing sentences into various types, which structure how ideas can relate to one another. Let us define a grammar structure as $G = (N, \Sigma, P, S)$ where  $N$ is a finite set of non-terminal symbols, $\Sigma$ is a finite set of terminal symbols (the actual words or tokens in the language), $P$ is a finite set of \textbf{production rules} that describe how non-terminals can be expanded into sequences of non-terminals and terminals and $S \in N$ is the \textbf{start symbol}, which we conventionally call \texttt{Sentence}.  Appendix \ref{sent} discusses the interplay of sentence and clauses in more detail.

  To understand the interplay of clauses and how they make up a sentence, we need to consider the types of sentences in English Language \cite{das2018novel}, which are:
\begin{itemize}
    \item Simple Sentences: Having only one independent clause and no dependent clause 
\end{itemize}
\[
S_{\text{simple}} = \{\, (NP,\, VP) \mid NP \in \mathcal{N},\; VP \in \mathcal{V} \}
\]
\begin{itemize}
    \item Complex Sentences: Having one independent clause and at least one dependent clause
    \[
S_{\text{complex}} = S_{\text{main}} \cup DC
\]
    \item Compound Sentences: Having two or more independent clauses joined by a conjunction and no dependent clause
    \[
S_{\text{compound}} = S_1 \oplus S_2 \text{ where } \oplus \text{ is a conjunction}
\] 
  \item Compound-Complex Sentences: Having two or more independent clauses joined by a conjunction and have at least one dependent clause
  \[
S_{\text{comp-comp}} = (S_1 \oplus S_2) \cup DC
\]
\end{itemize}
The core motivation behind this work stems from the assumption that LLMs emulate human reasoning \cite{wu2024thinking}. Additionally, as shown by Nurmalan \cite{hendrawati2018analysis}, human comprehension of sentence structure is far from uniform. Undergraduate students fail to accurately interpret CC sentences in 44.54\% of cases, followed by CD (23.2\%) and CX (22.13\%) sentences. In contrast, error rates drop significantly to 10.13\% for simple sentences. Notably, academic and scientific writing rarely employs simple sentences, favouring more elaborate constructions aligned with formal and jargon-heavy discourse. We posit that modelling and converting these complex sentence types into simpler forms enables more effective interpretation by LLMs, particularly for structured information extraction. 

A common misconception is that a simple sentence means a simplified or short sentence. However, as Phil \cite{atteberr_sentence_types} illustrates, even syntactically rich sentences, such as “Being an English teacher with a penchant for syntactical complexity, I love simple sentences upon getting up and before going to bed”—qualify as simple if they contain only one independent clause. Despite structural simplicity, such sentences may encode multiple relationships, contradicting the assumption that simple sentences yield only one extractable relation. Importantly, a simple sentence can feature compound subjects (“John and Mary run…”), compound predicates (“runs and jumps…”), or compound objects (“an apple and a banana…”). Our framework explicitly models these variations, ensuring RE remains robust across all syntactic permutations of the simple sentence form.

\subsection{Prompting Strategies}
\label{chpt2.1}
We explore four prompting strategies within our pipeline to evaluate their effectiveness in sentence restructuring and RE: GIP, FICL, CoT, and Hybrid CoT + FICL. GIP relies on general instructions without examples, offering baseline performance and broad generalizability \cite{ouyang2022training, kojima2022large}. FICL incorporates a small set of in-context examples to guide the model, improving structure-sensitive tasks like sentence decomposition \cite{brown2020language, li2023few}. COT prompting, which encourages step-by-step reasoning, has proven especially effective in multi-step reasoning and relation-rich generation \cite{wei2022chain, li2025structured}. Our hybrid CoT + FICL strategy combines the benefits of example-guided prompting with intermediate reasoning steps, significantly improving accuracy in sentence decomposition and RE. We benchmark each strategy across multiple subtasks in our pipeline and find that hybrid prompting consistently yields the most precise and coherent results, particularly in complex biomedical sentences, which is in tune with recent advances in prompt engineering that emphasize structure-aware and compositional prompting for complex NLP tasks \cite{kojima2022large}.

\section{Data}

\subsection{PubMed Lung Cancer Abstracts}
PubMed is an open biomedical literature repository. From PubMed, we randomly parsed 7,500 abstracts related to the lung cancer keyword, published between 2020 and 2025, to create our primary evaluation corpus. This dataset supports our co-reference resolution, sentence decomposition, and triple extraction tasks. The inclusion principle was any abstract that mentioned lung cancer, was open source and free to use, and we did not particularly exclude any research apart from those that fell outside the random sample. The data was sampled on March 18, 2025. The motivation behind using the lung cancer abstract was linked to a 20-year international study in Radiology \cite{henschke202320}. We believed that the dataset is well-versed and well-researched. 

\subsection{REBEL \cite{cabot2021rebel}}
We adopt the same 1,000-sample subset used in the EDC model \cite{zhang2024extract} for evaluation, originally drawn from the REBEL test partition of 105,516 entries, also published in EMNLP.

\subsection{WebNLG+2020 \cite{ferreira20202020}}
WebNLG+2020 (v3.0) is a semantic parsing benchmark containing text-triple pairs. We use its full test split of 1,165 samples covering 159 unique relation types.

\subsection{Wiki-NRE \cite{distiawan2019neural}}
Wiki-NRE is a distant supervision dataset for RE. We also used the same sample of 1,000 pairs used in the EDC model \cite{zhang2024extract}. The dataset contains 29,619 entries encompassing 45 distinct relation types
.
\subsection{CaRB \cite{bhardwajetal2019carb}}
The CaRB dataset is a benchmark for Open Information Extraction (OpenIE), created by re-annotating the original OIE2016 dataset with improved human judgments. The exact number of the final dataset is not known; what was reported in the paper was the devset from Amazon Mechanical Turk, which was 1,282 sentences. However, on the GitHub page, we found 50 unique sentences spanning through 172 lines. 

\begin{figure*}[htbp]
 \centering
 \includegraphics[width=1\textwidth]{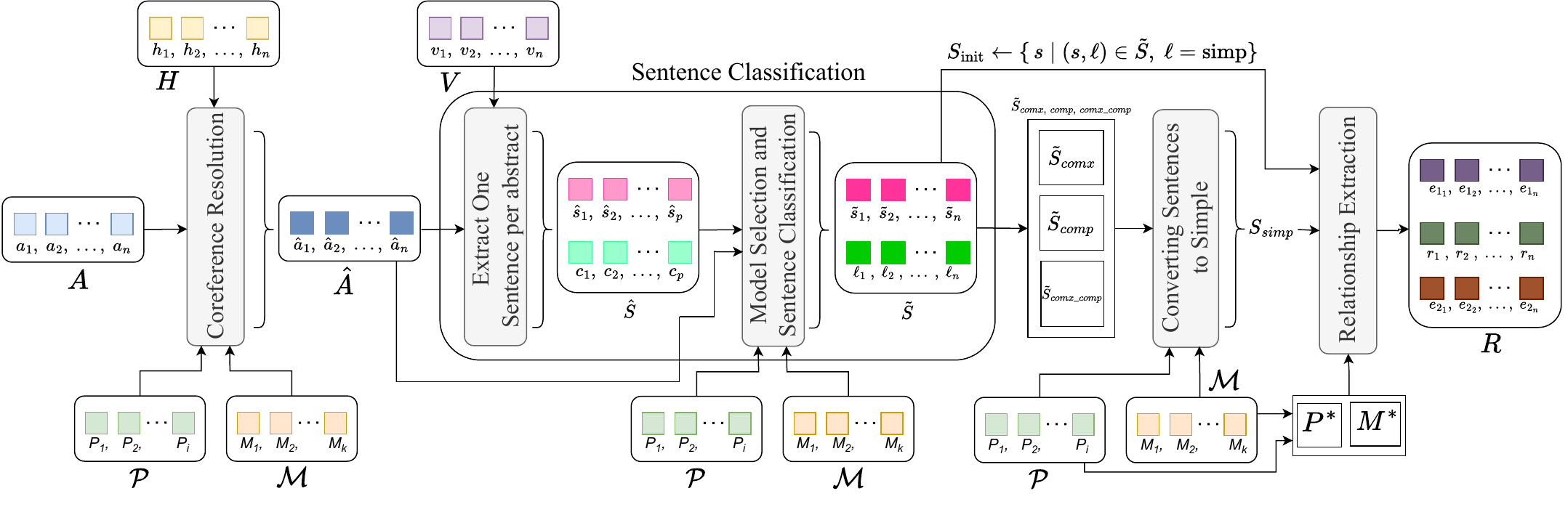}
 \caption{Overview of \textit{\textbf{CoDe-KG}}, the automated KG creation pipeline. First, the input set of abstracts $\absi$ is given to the \textbf{Coreference Resolution stage}. In this phase, a team of annotators $\annptators$, a collection of prompt strategies $\prompt$, and models $\model$ are jointly applied to produce the coreference‐resolved abstract set $\abso$, which is given as input in the \textbf{Sentence Classification stage}. With the help of verifiers $\ver$, prompting strategies $\prompt$ and models $\model$, a list of correctly classified sentences $\si$ with labels $\li$ is generated in this stage. Then, in the \textbf{Converting Sentences to Simple} stage, $\tilde S_{{comx},\; {comp},\; {comx\_comp}}$, prompt strategies $\prompt$, and models $\model$ are given as input and converted into simple sentences $\tilde S_{simp}$. In \textbf{Relationship Extraction} stage, $\tilde S_{simp}$, $S_{\mathrm{init}}$ and best model--prompt pair $(P^*,M^*)$ from previous stage are given as input and relationships ($\text{entity}_1$ $\eone$, $\text{relationship}$ $\rone$, $\text{entity}_2$ $\etwo$) are extracted for constructing KG.}
\label{fig:framework}
\vspace{-3mm}
\end{figure*}

\section{Methodology}
\label{sec:methodology}
In this research, we propose an automated KG creation pipeline, \textbf{\textit{CoDe-KG}}, for creating a KG from abstracts. Our approach, as shown in Figure~\ref{fig:framework}, consists of four key stages: doing coreference resolution, sentence classification, sentence conversion, and RE. In this section, we give a detailed overview of our pipeline implementation.

\subsection{Problem Setup}

Let the set of input abstracts be denoted by

\begin{equation}
A = \{\,a_1, a_2, \dots, a_n\},
\label{eq:abstracts}
\end{equation}

And let the set of valid relation triples extracted from \(A\) be denoted by

\begin{equation}
\begin{aligned}
R = \bigl\{(e_1, r, e_2)\mid {}&\;e_1, e_2 \in \mathcal{E},\\
&\;r \in \mathcal{R},\\
&\;(e_1, r, e_2)\text{ is valid}\bigr\}.
\end{aligned}
\label{eq:triples}
\end{equation}

Where \(\mathcal{E}\) is the set of all unique entities appearing in those triples and \(\mathcal{R}\) is the relation vocabulary. Let the resulting KG be denoted by

\begin{equation}
\mathcal{G} = (\mathcal{E}, \mathcal{R}),
\label{eq:graph}
\end{equation}

Our goal is to construct \(\mathcal{G}\) so that it faithfully represents all extracted factual relations across \(A\).

\subsection{Coreference Resolution}
In our proposed pipeline, the coreference‐resolution stage (as shown in Appendix: Algorithm~\ref{alg:coref_pipeline}) proceeds by creating a \textbf{gold-standard} through expert annotation, and then selecting the optimal \textbf{prompt-model combination} for creating coreference-resolved abstracts. First, we draw a random subset of size \(s\):
\[
  A' = \mathrm{UniformSample}(A,\,s).
\]
For four expert annotators--two with biological expertise and two with linguistic expertise--working in pairs to resolve coreference on each \(a\in A'\), producing
\[
  R_j(a) = f_{\mathrm{ann}}(h_j,\,a), 
  \quad j= 1, 2.
\]
Here, $f_{\mathrm{ann}}$ apply annotator $h_j$'s annotation procedure to abstract $a$, yielding the resolution $R_j(a)$.

We then define the gold set of abstracts $A$, which can be unanimously annotated:
\[
  G = \bigl\{\,a\in A' \;\bigm|\; 
    R_j(a)=R_k(a)\;\forall\,j,k
  \bigr\}.
\]
And extract the corresponding annotated gold standard
\[
  G' = \bigl\{R_j(a)\bigm|\;a\in G\bigr\},
\]
where any \(R_j(a)\) may be used since all agree.

Next, we exhaustively evaluate each prompt-model pair \((P,M)\in\mathcal P\times\mathcal M\) by generating predicted annotations on the gold inputs
\[
  \hat R_{P,M} = f_{\mathrm{prompt}}\bigl(P,\,M,\,G\bigr),
\]
and computing a score (e.g.,\ F\(_1\)) against the gold‐standard pairs:
\[
  S_{P,M} = \mathrm{score}\bigl(\hat R_{P,M},\,G'\bigr).
\]
We then select the best pair by
\[
  (P^*,M^*) = \mathrm{argmax}_{P\in\mathcal P,\;M\in\mathcal M} S_{P,M}.
\]
Finally, this optimal configuration is applied to the full collection:
\[
  \hat A = f_{\mathrm{prompt}}\bigl(P^*,\,M^*,\,A\bigr),
\]
yielding fully resolved co-reference annotations \(\hat A\). \\

This design ensures that (i) human expertise defines a robust gold standard through unanimous agreement, (ii) prompt-model selection is systematic and exhaustive, and (iii) large-scale annotation inherits the reliability established in the gold-standard phase.

\subsection{Sentence Classification}
The first step of the sentence classification stage (as shown in Appendix: Algorithm~\ref{alg:sample_and_extract}) processes the resolved abstracts \(\hat A\) by sampling per category, extracting one representative sentence from each sampled abstract, and keeping only those sentences on which two expert verifiers agree. For the five complexity categories
\[
  \mathcal{C} = \{\mathrm{simp},\mathrm{comx},\mathrm{comp},\mathrm{comx\_comp},\mathrm{incomp}\},
\]
we draw a random subset
\[
  A_c = \mathrm{UniformSample}(\hat A,\,p_c),
  \quad c\in\mathcal{C},
\]
so that \(\lvert A_c\rvert = p_c\). From each \(a\in A_c\) we then extract exactly one representative sentence:
\[
s = \begin{cases}
f_{\mathrm{create}}(\text{annotator},\,a), & \scriptstyle c\in\{\mathrm{simp},\mathrm{incomp}\},\\
f_{\mathrm{choose}}(a), & \text{otherwise}.
\end{cases}
\]
We aggregate all candidates into
\[
  S_{\mathrm{all}} = \bigcup_{c\in\mathcal{C}} S_c.
\]
Next, two expert verifiers \(v_1,v_2\) independently review every \(s\in S_{\mathrm{all}}\), and we keep only those sentence-category pairs on which they agree:

\[
\hat S = \left\{(s,c)\,\middle|\,
\begin{aligned}
  &c\in\mathcal{C},\;s\in S_c,\\
  &f_{\mathrm{ver}}(v_1,s)=f_{\mathrm{ver}}(v_2,s)
\end{aligned}
\right\}.
\]

The output \(\hat S\) is thus a high‐agreement, category‐labeled sentence set, ready to serve as the input for Step~2.

The second step of the sentence classification stage (as shown in Appendix: Algorithm~\ref{alg:full_sentence_classification}) takes the verified sentence-category set \(\hat S\) along with the set of prompting strategies \(\mathcal P\), and models \(\mathcal M\) as input to produce a fully labeled sentence corpus \(\tilde S\). First, we formed the training dataset
\[
  D = \{(s_i,y_i)\mid (s_i,y_i)\in\hat S\}.
\]
For each \(m\in\mathcal M\), train on \(D_{\mathrm{train}}\) and we computed its validation score
\[
  \mathrm{score}_m = \mathrm{Evaluate}(m,\,D_{\mathrm{val}}).
\]
Then we selected the best model
\[
  m^* = \mathrm{argmax}_{m\in\mathcal M} \mathrm{score}_m.
\]
Finally, we applied \(m^*\) to every sentence in the fully coreference‐resolved abstract set \(\hat A\):
\[
  \tilde S = \bigl\{(s,\ell_s)\mid s\in\mathrm{Sentences}(\hat A),\;\ell_s = m^*(s)\bigr\}.
\]
The resulting \(\tilde S\) is the complete collection of sentence-label pairs for downstream tasks.

\subsection{Converting Sentences to Simple}


The approach of this stage (as shown in Appendix: Algorithm~\ref{alg:unified_simplify}) consists of prompt-model selection for each category, and large‐scale sentence simplification using the selected configurations.

For each category \(c\), we hold out the set \(S_c\) of complex (\texttt{comx}), compound (\texttt{comp}), and complex-compound (\texttt{comx\_comp}) sentences, and exhaustively evaluate every prompt-model combination \((P,M)\in\mathcal P\times\mathcal M\). We compute a performance score via
\newcommand{\EPM}[3]{\mathrm{EvaluatePromptModel}(#1,#2,#3)}
\[
  \mathrm{score}_{P,M}(c) = \EPM{P}{M}{S_c}.
\]

and choose
\[
  (P^*_c,\,M^*_c) = \mathrm{argmax}_{(P,M)} \mathrm{score}_{P,M}(c).
\]

With \((P^*_c,M^*_c)\) fixed for each category, we process every sentence \(s\in S_c\) by invoking:
\[
  \hat s = f_{\mathrm{prompt}}\bigl(P^*_c,\,M^*_c,\,s\bigr),
  \quad
  S_{\mathrm{simp}} \leftarrow S_{\mathrm{simp}} \cup \{\hat s\}.
\]
Once all categories are processed, \(S_{\mathrm{simp}}\) constitutes the collection of simplified sentences from complex, compound, and complex-compound sentences.

\subsection{Relationship Extraction}
In this stage, we implement RE (as shown in Appendix: Algorithm~\ref{alg:relation_extraction}) through \textbf{sentence consolidation} and \textbf{triple generation}. First, we form the working sentence set by combining:
\[
\begin{split}
  S_{\mathrm{init}}
    &= \{\,s \mid (s,\ell)\in\tilde S,\;\ell=\mathrm{simp}\},\\
  S
    &= S_{\mathrm{simp}} \;\cup\; S_{\mathrm{init}}.
\end{split}
\]

Here, \(S_{\mathrm{simp}}\) is the set of all sentences produced by the simplification stage, while \(S_{\mathrm{init}}\) contains those initially classified as simple. Next, for each \(s\in S\), we extract a candidate relation triple via the $f_{\mathrm{rel}}(s)$ function, where the best prompt and model combination $(P^*,\;M^*)$ from the previous stage was used. Here,
\[
  (e_1,\,r,\,e_2) \;=\; f_{\mathrm{rel}}(P^*,M^*,s).
\]
We collect only non‐empty outputs:
\[
  R \;\leftarrow\; R \cup \{(e_1,r,e_2)\}
  \quad\text{if}\quad (e_1,r,e_2)\neq \varnothing.
\]
Upon completion, \(R\) holds all valid \((\text{entity}_1,\,\text{relation},\,\text{entity}_2)\) triples. Finally, we assemble the extracted triples into our knowledge graph. Let
\[
  \mathcal{E} \;=\; \bigcup_{(e_1,r,e_2)\in R}\{e_1,e_2\},
  \quad 
  \mathcal{R} \;=\; \bigcup_{(e_1,r,e_2)\in R}\{r\},
\]
and define the graph as
\[
  \mathcal{G} \;=\; (\mathcal{E},\,\mathcal{R}).
\]
Each triple \((e_1,r,e_2)\in R\) becomes a directed, labeled edge from node \(e_1\) to node \(e_2\).  This knowledge graph \(\mathcal{G}\) now encodes all valid factual relations extracted across the corpus and can be used for downstream querying and inference.

\section{Experiments}
\label{sec:exp}

\begin{table*}[t]
\centering
\small
\begin{tabular}{l l r r r r}
\toprule
\textbf{Model} & \textbf{Prompt} & \textbf{MUC (\%)} & \textbf{B$^3$ (\%)} & \textbf{CEAF (\%)} & \textbf{CoNLL (\%)} \\
\midrule
\textbf{Mixtral-8x7B-Instruct-v0.1}   & FICL      & 32.42 & 70.61 & 70.61 & \textbf{57.88} \\
Llama-3.1-8B-Instruct        & FICL      & 27.16 & 69.57 & 69.57 & 55.43 \\
Llama-3.2-3B-Instruct        & COT\_FICL & 16.98 & 70.7  & 70.7  & 52.79 \\
Llama-3.3-70B-Instruct       & FICL      & 31.25 & 70.94 & 70.94 & 57.71 \\
Mistral-7B-Instruct-v0.3     & COT\_FICL & 18.58 & 70.74 & 70.74 & 53.35 \\
\bottomrule
\end{tabular}
\caption{Comparison of F$_1$ Scores by Models.}
\label{tab:f1-comparison}
\end{table*}

\subsection{Experiment 1: Results of Co-reference Resolution}
Biomedical text is harder to understand and therefore, RE is perceived to be harder \cite{johnson2023biomedical}. Therefore, we construct our benchmark dataset of 190 coreference abstracts in biomedical literature on Lung Cancer to evaluate the performance of LLM. Therefore, we crafted the SOTA prompts discussed in chapter \ref{chpt2.1}. The prompts we finally used were the COT+FICL prompt after experimenting on the different prompts, which are in Appendix \ref{app:corefprompt}. We then randomly sampled 190 abstracts from the full set of 7,500. Each abstract was independently annotated by two domain experts and two language experts. After the initial pass, the experts exchanged annotations and discussed any discrepancies. Full annotation information available in Appendix \ref{app:corefannotation}.


We evaluated several LLMs on our benchmark. Results were poor for most models. Predictions were scored using MUC, B$^{3}$, CEAF$_{4}$, and the CoNLL F$_1$ aggregate \cite{pradhan2012conll}. Deepseek‐distill‐Qwen‐7B, Qwen‐Chat‐7B, and Qwen‐7B scored 0\% F$_1$. Deepseek‐7B, Deepseek‐6.7B, and Deepseek‐Prover‐7B scored below 2\% F$_1$. Deepseek‐distill‐Llama‐8B scored below 10\% F$_1$ in all categories. Table~\ref{tab:f1-comparison} lists F$_1$ scores for the models doing co-reference resolution and was benchmarked with ChatGPT o4-mini-high and ChatGPT-4.5 responses as a baseline. ChatGPT o4-mini-high did the best overall with an F$_1$ of approximately 63\% using the FICL prompt. We bolded the best open-source model, which was comparable to the closed source models for this task.

The evaluation of the co-reference was done using a cosine similarity score of 0.9, because we noticed that 99\% of all values that were marked at 0.9 correctly but not exact match were actually correct, just differing in preposition. For instance, the gold standard says "a house", and the model says "house". At 0.8\% the values were not significant to be considered, therefore, we stuck to using a cosine similarity score of 0.9 for the co-reference evaluation. 




\subsection{Experiment 2: Syntactic Sentence Classification}
We created a dataset for classification, and all details are given in Appendix \ref{app:sentence}. We fine‐tuned six transformer‐based models and two smaller LLMs on the training set and evaluated them on the test set. Table~\ref{tab:exp2-results} reports test accuracy and macro‐averaged F$_1$ for each model. We evaluated the entire test set on a GPT-4o model.

\begin{table}[ht]
\centering
\caption{Sentence‐type classification results (train set: 2,00- sentences test set: 5,269 sentences)}
\label{tab:exp2-results}
\begin{tabular}{lcc}
\toprule
Model             & Accuracy & F$_1$$_{\mathrm{macro}}$ \\ 
\midrule
BERT              & 87.25\%  & 86.14\%                  \\ 
BERT‐Large        & \textbf{87.68\%}  & \textbf{86.69\%}       \\ 
BioBERT           & 87.19\%  & 86.21\%                  \\ 
BioBERT‐Large     & 85.97\%  & 85.16\%                  \\ 
ClinicalBERT      & 86.92\%  & 85.87\%                  \\ 
RoBERTa           & 87.13\%  & 85.83\%                  \\ 
Gemma3 1-B        & 9.24\%   & 4.15 \%                  \\
LLama3.2 1-B      & 17.71 \% & 0.27 \%                  \\
\textbf{GPT 4-0}       & 80.30 \% & 76.14\%                  \\
Random Guessing Lowest & 8.83 \% & 3.24\%                \\
Random Guessing Highest & 32.61 \% & 9.84 \%              \\
\bottomrule
\end{tabular}
\end{table}

\subsection{Experiment 3: Evaluating the Prompting Strategies and Semantic Conversion}
\begin{table}[ht]
  \centering
  \setlength{\tabcolsep}{4pt}
  \scriptsize
  \caption{Model performance on the conversion of \textit{Compound to Simple Sentences}}
  \begin{tabular}{llccc}
    \toprule
    \textbf{Model} & \textbf{Macro Avg.} & \textbf{Exact-Match} & \textbf{RMSE} \\
    \midrule
    DeepSeek-LLM-67B & 15.56\% & 14.67\% & 1.5891 \\
    DeepSeek-LLM-7B & 55.83\% & 50.00\% & 1.1506 \\
    DeepSeek-R1-Distill-Llama-8B & 68.04\% & 65.00\% & 1.0571 \\
    DeepSeek-Prover-V1.5-7B & 81.38\% & 76.33\% & 0.8591 \\
    Llama-3.3-70B & 95.30\% & 81.00\% & 0.3213 \\
    \textbf{Llama-3-8B} & \textbf{99.78\%} & \textbf{98.00\%} & \textbf{0.1078} \\
    Mistral-7B-Instruct-v0.3 & 96.64\% & 90.33\% & 0.2356 \\
    Mixtral-8x7B-Instruct-v0.1 & 96.14\% & 91.67\% & 0.2323 \\
    Qwen-7B & 56.17\% & 54.00\% & 1.2339 \\
    Qwen-7B-Chat & 1.33\% & 1.33\% & 1.7193 \\
    \textit{GPT 4 o} & \textit{91.68\%} & \textit{77.67\%} & \textit{0.4204} \\
    \textit{GPT 4 o-3} & \textit{87.69\%} & \textit{68.67\%} & \textit{0.5924} \\
    \bottomrule
  \end{tabular}
\end{table}

\begin{table}[ht]
  \centering
  \setlength{\tabcolsep}{4pt}
  \scriptsize
  \caption{Model performance on the conversion of \textit{Complex to Simple Sentences}}
  \begin{tabular}{llccc}
    \toprule
    \textbf{Model} & \textbf{Macro Avg.} & \textbf{Exact-Match} & \textbf{RMSE} \\
    \midrule
    DeepSeek-LLM-67B & 26.23\% & 22.00\% & 1.7562 \\
    DeepSeek-LLM-7B & 63.30\% & 62.33\% & 1.1866 \\
    DeepSeek-R1-Distill-Llama-8B & 43.78\% & 33.00\% & 1.5540 \\
    DeepSeek-Prover-V1.5-7B & 91.33\% & 65.67\% & 0.8841 \\
    \textbf{Llama-3.3-70B} & \textbf{99.59\%} & \textbf{98.67\%} & \textbf{0.1364} \\
    Llama-3-8B & 97.17\% & 92.67\% & 0.2866 \\
    Mistral-7B-Instruct-v0.3 & 93.73\% & 81.00\% & 0.3114 \\
    Mixtral-8x7B-Instruct-v0.1 & 98.48\% & 94.67\% & 0.1533 \\
    Qwen 7B & 64.61\% & 64.00\% & 1.1987 \\
    Qwen-7B-Chat& 9.24\% & 7.67\% & 1.8699 \\
    \textit{GPT 4 o} & \textit{96.72\%} & \textit{91.33\%} & \textit{0.3050} \\
    {\textit{GPT 4 o-3}} & \textit{99.05\%} & \textit{97.00\%} & \textit{0.1714} \\
    \bottomrule
  \end{tabular}
\end{table}

\begin{table}[ht]
  \centering
  \scriptsize
  \setlength{\tabcolsep}{4pt}
  \caption{Model performance on the conversion of \textit{Compound-Complex to Simple Sentences}}
  \begin{tabular}{llccc}
    \toprule
    \textbf{Model} & \textbf{Macro Avg.} & \textbf{Exact-Match} & \textbf{RMSE} \\
    \midrule
    DeepSeek-LLM-67B & 28.42\% & 11.00\% & 1.6178 \\
    DeepSeek-LLM-7B & 49.45\% & 32.33\% & 1.2848 \\
    DeepSeek-R1-Distill-Llama-8B & 37.25\% & 21.67\% & 1.5183 \\
    DeepSeek-Prover-V1.5-7B & 71.12\% & 51.00\% & 0.8411 \\
    Llama-3.3-70B & 89.57\% & 72.67\% & 0.4836 \\
    Llama-3-8B & 91.71\% & 78.00\% & 0.4544 \\
    Mistral-7B-Instruct-v0.3 & 91.19\% & 80.00\% & 0.3273 \\
    \textbf{Mixtral-8x7B-Instruct-v0.1} & \textbf{92.16\%} & 76.67\% & \textbf{0.2727} \\
    Qwen 7B & 56.68\% & 40.33\% & 1.1977 \\
    Qwen-7B-Chat& 15.81\% & 4.33\% & 1.7556 \\
    \textit{GPT 4 o} & \textit{82.75\%} & \textit{68.33\%} & \textit{0.5354} \\
    \textit{GPT 4 o-3} & \textit{94.10\%} & \textbf{\textit{81.67\%}} & \textit{0.2320} \\
    \bottomrule
  \end{tabular}
\end{table}

We created a systematic structure (see Appendices \ref{app:cx}, \ref{app:cd}, and \ref{app:cc} for the systemic conversion process) of evaluating how a model would convert a cx, cd or cc sentence to a simple one, thereby, evaluating the four prompting strategies--GIP, FICL, COT, and COT+FICL prompts (see Appendix~\ref{app:cot+ficl}).  From the 65,175 sentences extracted from 7,500 abstracts, our classifier labelled 42,282 as complex, 4,942 as compound, 2,366 as compound-complex, 13,465 as simple, and 2,120 as incomplete. We then randomly sampled 300 sentences each from the complex, compound, and compound-complex classes and translated them into simple sentences. These sample sizes at 300 achieve 95\% confidence for their respective populations with margins of error of $\pm5.62\%$, $\pm5.57\%$, and $\pm5.26\%$, respectively. We tested on the top performing models and evaluated their performance on the different prompting strategies. The results are shown in Table \ref{tab:prompt_performance}, with the hybrid prompt performing the best in all cases.

\subsection{Extracting Relationship Pairs from Simple Sentences}
With the total number of simple sentences exceeding 177,000, we used a carefully crafted COT + FICL prompt as it has shown from data to perform the best. The annotators AB and CD studied 100 sentences generated from each model that were tested below and came to an agreement that the Mixtral-8x7B-Instruct-v0.1 model performed well with a 99\% accuracy in parsing relationships from simple sentences, which also involved capturing multiple relationships in text. It was a Boolean task. The table for the task is Table \ref{tab:prompting-eval4} located in Appendix \ref{app:t3}.

\section{Discussion}
\label{sec:results}
We evaluated the results on a subset of 200 samples each from Rebel, Web-NLG, and Wiki-NRE obtained from the GitHub page of \cite{zhang2024extract}. We also evaluated the 50 unique sentences from the CaRB dataset \cite{bhardwajetal2019carb}. 

\begin{table}[ht]
  \centering
  \scriptsize
  \caption{Evaluation metrics across benchmarks}
  \begin{tabular}{lcccc}
    \toprule
    \textbf{Metrics} & \textbf{ReBEL} & \textbf{Web-NLG2} & \textbf{Wiki-NRE} & \textbf{CaRB} \\
    \midrule
    Exact-Match           & 14.50\% & 43.00\% & 7.00\%  & 43.14\% \\
    Prec Macro            & 72.97\% & 80.70\% & 60.26\% & 66.96\% \\
    Rec Macro             & 59.88\% & 71.64\% & 60.13\% & 68.43\% \\
    F1-Score Macro        & 65.78\% & 75.90\% & 60.20\% & 62.61\% \\
    Prec Micro            & 66.34\% & 79.35\% & 59.02\% & 63.69\% \\
    Rec Micro             & 59.88\% & 72.32\% & 58.67\% & 59.52\% \\
    F1-Score Micro        & 62.94\% & 75.67\% & 58.84\% & 61.54\% \\
    RMSE                   & 0.8813  & 0.5785  & 0.9648  & 0.3633  \\
    \bottomrule
  \end{tabular}
  \label{tab:eval_metrics}
\end{table}

The results show that even with the extra aid, LLMs still have difficulty parsing relationships the way humans can. Comparing to current SOTA techniques, for the ReBEL benchmark, our model achieves a macro-F1 of 65.78\% and a micro-F1 of 62.94\%, surpassing the published macro-F1 of 51.0\% \cite{cabot2021rebel} but falling short of the SOTA micro-F1 of 74.0\% \cite{cabot2021rebel}, which tells us that our pipeline can handle diverse, low-frequency relations reasonably well (macro), yet would not perform the best on the most common triplets that dominate micro averaging. The very low exact-match rate (14.5\%) are due to the sentence decomposition. In  Web-NLG2 and Wiki-NRE RE, we record a micro-F1 of  75.67\% and 58.84\% respectively, which falls largely behind the SOTA at 93.6\% \cite{ferreira20202020} and GenIE’s 91.48\% \cite{distiawan2019neural} respectively. However, knowing that this generative task without any fine-tuning shows opportunities for great improvement. In the CaRB OpenIE benchmark, we achieve micro-F1 61.54\% and macro-F1 62.61\%, compared to DetIE’s SOTA 67.7\% \cite{bhardwajetal2019carb}. 

Though the results were not as high as we expected, they look promising for an unsupervised approach. Future work would look into fine-tuning LLMs for specific tasks like these.  To justify the need for our pipeline, we performed an ablation study on 23 articles already in our coreferenced set.  Table \ref{tab:triple_results}, Appendix \ref{app:ablation}, shows that removing coreference resolution or sentence-level decomposition hurts our performance sharply, and without these decomposition, our recall falls below half, which explains to us that biomedical sentences have a lot of nuanced co-referent names that must be split before extraction. Therefore, having this pipeline is a reliable way to use LLMs for generated content. We have baselines like ChatGPT-4.5, which do have near-perfect precision. They often summarize relationships and do not give the user the ability to pick and choose, as it makes the decision on the user's behalf. The DeepSeek-405B model performs well too, as it has a higher recall, but introduces a lot of errors. It is interesting to see how our pipeline maintained high precision and recall, but extracted some relationships beyond human reference. The results were hand-graded by humans using the gold standard. 

\subsection{Error Analysis}
From the analysis, we classified the errors into 3 types, namely: "missing", "spurious", and "relation-mismatch". Missing error happens when a relationship type is not considered. Spurious is non-existent but invented relationships and relation mismatch occurs when there is a switch between Entity 1 and Entity 2 or the relationship actually exists with another Entity (Entity3), but the model misclassified it. Furthermore, in the Rebel dataset, the relation-extraction is consistent, as it only contains four variables across the dataset. We noticed that the pipeline parses some “less important” or “missed relationships”, which we categorized as spurious in this evaluation. During the evaluation, if a model outputs something not in the gold, but is a valid relationship from the text, the model is not given a score (neither penalized nor awarded). 

\begin{figure}[ht]
  \centering
  \includegraphics[width=0.98\linewidth]{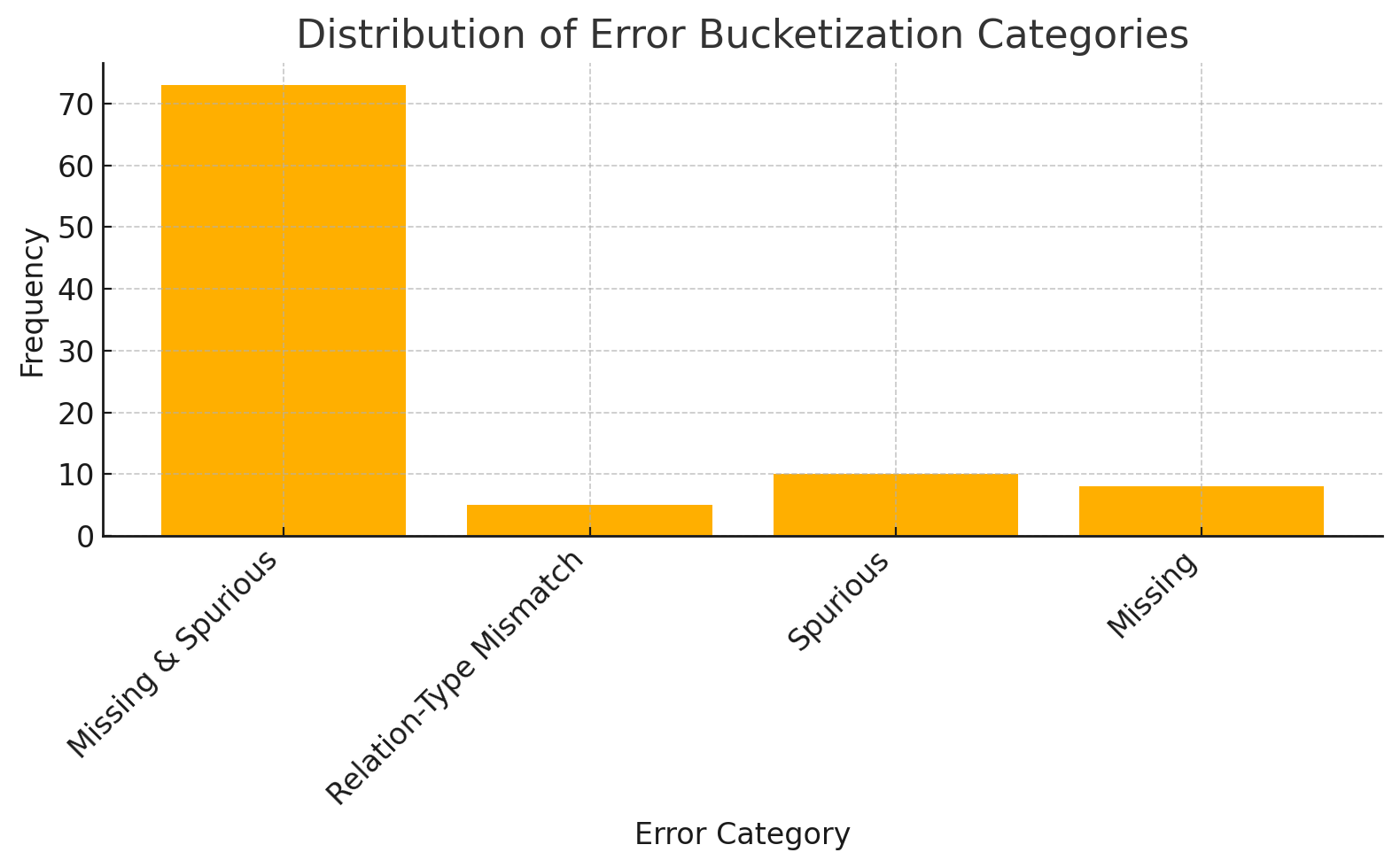}
  \caption{Distribution of error bucketization categories.}
  \label{fig:error}
\end{figure}

However, if it outputs something not in the gold, and cannot be inferred from the text, then we penalize the model. We also noticed a few relation-type mismatches, where the model defaults to a more simplistic relationship type, like “is”, “was”, rather than focusing on the actual relationship. For example, the model used a generic “is” for party membership instead of “member of political party”, in the sentence “Anju Dhillon (born 1979) is a Canadian Liberal politician, who was elected to represent the riding of Dorval-Lachine-LaSalle in the House of Commons of Canada in the 2015 federal election.”. Missing and Spurious relationships amongst the errors were profound, as the model typically highlights 3 out of 4 of the relationship types in REBEL and typically omits one. A combination of spurious and missing relationships characterized 73\% of the errors, while spurious relationships alone scored 11\%, and missing relationships 9\% and relation-type is 7\%.  Interestingly, more often than not, it highlights newer relationships that we were not considering.

\section{Conclusion}
\label{sec:conc}
In this work, we created a pipeline and verified that using sentence decomposition on open-source models actually helps the model think through each problem uniquely. By releasing our implementation, annotated datasets, and evaluation scripts, we aim to promote reproducibility and accelerate future work in information extraction and co reference resolution. The annotated datasets from humans include 190 abstract texts, 7248 rows for sentence classification, and 900 rows for sentence decomposition.


\section*{Limitations}
\label{sec:limitations}

The limitations of our paper are as follows:

\begin{itemize}
    \item Before, our pipeline would be of industry-standard, it is imperative that we find a solution to coreference resolution and its weaknesses. Future work would look into how to improve coreference resolution in thick abstract texts.

    \item Using just open-source models, though comparative with ChatGPT tends to fail at times, and thereby, average the RE score. Future iterations of the work would employ agent systems that can scan for missing or inconsistent data through a feedback loop and judge the quality to improve the output. 

    \item We only focused on prompting this time; future iterations would look at fine-tuning strategies like Parameter-Efficient Fine-tuning (PEFT) that could help the LLM perform better in weak tasks.

    \item We had a lot of human annotations for quality. Future work would engage more code scripts, so we can evaluate a wider range of output. 

    \item We acknowledge that our pipeline is costly computationally, and as such, it might not be possible to run with low-level resources, and though the CoT+FICL prompting did well, it might incur a non-trivial inference time and token consumption. However, since this is a one-off knowledge graph, it is still deployable.

    \item Our evaluation is confined to scientific and benchmark datasets. We have not tested the pipeline on domains such as newswire, legal text, or social media. Thus, its generalizability to noisier or less formal text remains unverified.  
\end{itemize}

\section*{Acknowledgements}
We thank the expert annotators for their work. The biologist: Thelma Emeji and Olaniyi Glory; and the linguist: Oluwafaratunmi Olakanmi,  and Godman Adebayo. We also thank the rest of the other annotators who did not want their names mentioned. All instructions given to the annotators were the same as the prompt. Each annotator was paid weekly at a rate of 40,000 Nigerian Naira, over a period of 8 weeks, and this project was self-funded.

\bibliography{ref}

\appendix

\appendix
\section{Sentence Steps} \label{sent}
Sentences are made up of clauses, phrases, words, and punctuation. Phrases are a group of words that act as a single unit but do not have both a subject and a predicate. The common types are noun phrase $NP$, verb phrase $VP$, and prepositional phrases $PP$. A sentence in English is defined as: $Sentence\rightarrow NP, VP$. This means every sentence must have a noun phrase, typically including the subject and the verb. Not all sentences have a predicate or an object; however, this is very uncommon in academic writing. Noun phrases are defined as: $NP \rightarrow$ $|$Det $N$ $|$ Det Adj $N$ $|$ Pronouns $|$Proper Nouns$|$, while verb phrases are defined as: $VP \rightarrow | V | V NP | V NP PP|$, where $Det$ is a determiner: for example, "a", "the", "an", "those", etc., $Adj$ is an adjective, and $V$ is a verb. Clauses are typically of two types: independent clauses (IC) and dependent clauses (DC). An IC can stand alone as a complete sentence, while the DC relies on an IC. Dependent clauses start with subordinating conjunctions like “because”, “although”, “when”, “if”, etc., and are then followed by an IC to make a complete sentence. $Clause\rightarrow NP, VP$. Mathematically, it is represented by  $DC \rightarrow$ $Subconj$ $IC$.

\section{Algorithms}

\begin{algorithm}[tb]
\caption{Coreference Resolution}
\label{alg:coref_pipeline}
\textbf{Input}: abstracts $A$, subset size $s$, annotators $H$, prompt strategies $\mathcal{P}$, models $\mathcal{M}$\\
\textbf{Output}: resolved abstracts $\hat A$
\begin{algorithmic}[1]
    \State $A' \gets \mathrm{UniformSample}(A,\,s)$  
      \newline\Comment{Select $s$ abstracts at random}
    \ForAll{$a \in A'$}
        \ForAll{$h_j \in H$}
            \State $R_j(a) \gets f_{\mathrm{ann}}(h_j,\,a)$  
              \newline\Comment{Annotator $h_j$ resolves coreference on $a$}
        \EndFor
    \EndFor 
    \State $G \gets \{\,a\in A' \mid R_j(a)=R_k(a)\;\forall\,j,k\}$ \newline\Comment{Gold set of unanimous abstracts}
    \State $G' \gets \{R_j(a) \mid a\in G\}$  
    \newline\Comment{Gold‐standard annotated abstracts}
    \State $(P^*,M^*) \gets (\varnothing,\varnothing)$; $\mathrm{bestScore}\gets -\infty$
    \ForAll{$P \in \mathcal{P}$}
        \ForAll{$M \in \mathcal{M}$}
            \State $\hat R_{P,M} \gets f_{\mathrm{prompt}}(P,\,M,\,G)$ 
              \newline\Comment{Predict annotations on $G$}
            \State $S_{P,M} \gets \mathrm{score}(\hat R_{P,M},\,G')$  
              \newline\Comment{Evaluate predictions against $G'$}
            \If{$S_{P,M} > \mathrm{bestScore}$}
                \State $(P^*,M^*) \gets (P,M)$  
                \State $\mathrm{bestScore} \gets S_{P,M}$  
                  \newline\Comment{Update best prompt-model pair}
            \EndIf
        \EndFor
    \EndFor
    \State $\hat A \gets f_{\mathrm{prompt}}(P^*,\,M^*,\,A)$  
      \newline\Comment{Resolve coreference on full collection}
    \State \Return $\hat A$
\end{algorithmic}
\end{algorithm}

\begin{algorithm}[tb]
\caption{Step 1: Sample Abstracts, Extract One Sentence, and Verify}
\label{alg:sample_and_extract}
\begin{algorithmic}[1]
    \State \textbf{Input:} resolved abstracts $\hat A$, sample sizes $p_{\mathrm{simp}}, p_{\mathrm{comx}}, p_{\mathrm{comp}}, p_{\mathrm{comx\_comp}}, p_{\mathrm{incomp}}$, expert verifiers $V=\{v_1,v_2\}$
    \State \textbf{Output:} verified sentences with category labels $\hat S$
    \ForAll{category $c$ in \{\texttt{simp}, \texttt{comx}, \texttt{comp}, \texttt{comx\_comp}, \texttt{incomp}\}}
        \State $A_c\gets\mathrm{UniformSample}(\hat A,\,p_c)$  
        \Comment{Select $p_c$ abstracts for $c$}
        \State $S_c\gets\emptyset$
        \ForAll{$a\in A_c$}
            \If{$c\in\{\mathrm{simp},\mathrm{incomp}\}$}
                \State $s\gets f_{\mathrm{create}}(\text{annotator},a)$  
                  \Comment{Create sentence for simple/incomplete}
            \Else
                \State $s\gets f_{\mathrm{choose}}(a)$  
                  \Comment{Choose sentence for other categories}
            \EndIf
            \State $S_c\gets S_c\cup\{s\}$
            \Comment{Collect one sentence per abstract}
        \EndFor
    \EndFor
    \State \textbf{Ensure}\;$\bigcup_cA_c=\hat A$  
    \Comment{Coverage of all abstracts}
    \State $S_{\mathrm{all}}\gets\bigcup_cS_c$  
    \Comment{All candidate sentences}
    \State $\hat S = \{(s,c)\,\mid\,c\in\mathcal C,\;s\in S_c,\;f_{\mathrm{ver}}(v_1,s)=f_{\mathrm{ver}}(v_2,s)\}$
    \Comment{Keep only unanimously verified sentences with their category; where $\mathcal C=\{\mathrm{simp},\mathrm{comx},\mathrm{comp},\mathrm{comx\_comp},\mathrm{incomp}\}$}
    \State \Return $\hat S$
\end{algorithmic}
\end{algorithm}

\begin{algorithm}[tb]
\caption{Step 2: Model Selection and Full Classification}
\label{alg:full_sentence_classification}
\begin{algorithmic}[1]
    \State \textbf{Input:} verified dataset 
      $D = \{(s_i,y_i)\mid s_i\in\hat S\}$, 
      candidate models $\mathcal M$, 
      resolved abstracts $\hat A$
    \State \textbf{Output:} full classification 
      $\tilde S = \{(s,\ell)\mid s\in\mathrm{Sentences}(\hat A)\}$
    \State $\mathrm{bestScore}\gets -\infty$, \quad $m^*\gets \varnothing$
    \ForAll{$m \in \mathcal M$}
        \State Train $m$ on training split of $D$
        \State $\mathrm{score}\gets \mathrm{Evaluate}(m,\text{val split of }D)$
        \If{$\mathrm{score}>\mathrm{bestScore}$}
            \State $\mathrm{bestScore}\gets \mathrm{score}$
            \State $m^*\gets m$
        \EndIf
    \EndFor
    \State $\tilde S \gets \emptyset$
    \ForAll{abstract $a \in \hat A$}
        \ForAll{sentence $s \in \mathrm{Sentences}(a)$}
            \State $\ell \gets m^*.\mathrm{classify}(s)$  
            \Comment{Classify each sentence in the main abstracts}
            \State $\tilde S \gets \tilde S \cup \{(s,\ell)\}$
        \EndFor
    \EndFor
    \State \Return $\tilde S$
\end{algorithmic}
\end{algorithm}

\begin{algorithm}[tb]
\caption{Unified Sentence Simplification}
\label{alg:unified_simplify}
\begin{algorithmic}[1]
    \State \textbf{Input:} sentence sets $\{S_c\}_{c\in\{\mathrm{comx},\mathrm{comp},\mathrm{comx\_comp}\}}$, prompt strategies $\mathcal P$, models $\mathcal M$
    \State \textbf{Output:} simplified sentences $S_{\mathrm{simp}}$
    \State $S_{\mathrm{simp}}\gets\emptyset$  \newline\Comment{Initialize output set}
    \ForAll{category $c\in\{\mathrm{comx},\mathrm{comp},\mathrm{comx\_comp}\}$}  \newline\Comment{Iterate over sentence categories}
        \State $\mathrm{bestScore}\gets -\infty$  \newline\Comment{Reset best score}
        \ForAll{$P\in\mathcal P$}  
        \newline\Comment{For each prompting strategy}
            \ForAll{$M\in\mathcal M$}
            \newline\Comment{For each model}
                \State $\mathrm{score}\gets \mathrm{EvaluatePromptModel}(P,M,S_c)$  \newline\Comment{Evaluate on $S_c$}
                \If{$\mathrm{score}>\mathrm{bestScore}$}
                    \State $\mathrm{bestScore}\gets\mathrm{score}$; $(P^*,M^*)\gets(P,M)$  \newline\Comment{Update best pair}
                \EndIf
            \EndFor
        \EndFor
        \ForAll{$s\in S_c$}  
        \newline\Comment{Simplify each sentence in category $c$}
            \State $\hat s\gets f_{\mathrm{prompt}}(P^*,M^*,s)$  
            \newline\Comment{Generate simplified sentence}
            \State $S_{\mathrm{simp}}\gets S_{\mathrm{simp}}\cup\{\hat s\}$  \newline\Comment{Collect simplified sentence}
        \EndFor
    \EndFor
    \State \Return $S_{\mathrm{simp}}$  
    \newline\Comment{Return all simplified sentences}
\end{algorithmic}
\end{algorithm}

\begin{algorithm}[tb]
\caption{Relationship Extraction from Simplified Sentences}
\label{alg:relation_extraction}
\begin{algorithmic}[1]
    \State \textbf{Input:} 
      simplified sentences $S_{\mathrm{simp}}$, 
      classified sentences $\tilde S = \{(s,\ell)\}$
      best prompting strategy-model pair $(P^*,M^*)$
    \State \textbf{Output:} relation triples 
      $R = \{(e_1,r,e_2)\}$
    \medskip
    \State $S_{\mathrm{init}} \gets \{\,s \mid (s,\ell)\in\tilde S,\;\ell=\mathrm{simp}\}$  
      \Comment{Select only initially classified simple sentences}
    \State $S \gets S_{\mathrm{simp}}\;\cup\;S_{\mathrm{init}}$  
      \Comment{Combine with previously simplified sentences}
    \State $R \gets \emptyset$  \Comment{Initialize relation set}
    \ForAll{$s \in S$}  
        \State $(e_1,r,e_2) \gets f_{\mathrm{rel}}(P^*,M^*,s)$  
          \Comment{Extract (entity$\mathrm{_1}$, relationship, entity$\mathrm{_2}$)}
        \If{$(e_1,r,e_2)\neq \varnothing$}
            \State $R \gets R \cup \{(e_1,r,e_2)\}$  
            \Comment{Keep valid triples}
        \EndIf
    \EndFor
    \State \Return $R$  
      \Comment{All extracted relation triples}
\end{algorithmic}
\end{algorithm}

\onecolumn

\section{Tables}
\subsection{Coreference Resolution}
\begin{table*}[ht]
  \centering
  \setlength{\tabcolsep}{1.5pt}
  \scriptsize
  \caption{Coreference resolution performance using cosine similarity across models and prompting styles.}
  \label{tab:coref_performance}
  \begin{tabular}{llcccc}
    \toprule
    \textbf{Model} & \textbf{Prompt} & \textbf{MUC (\%)} & \textbf{B3 (\%)} & \textbf{CEAF (\%)} & \textbf{CoNLL (\%)} \\
    \midrule
    \multirow{4}{*}{\textbf{Mixtral-8x7B-Instruct-v0.1}}
      & GIP           & 13.34 & 70.96 & 70.96 & 51.75 \\
      & COT           & 14.50 & 70.87 & 70.87 & 52.08 \\
      & \textbf{FICL}          & \textbf{32.42} & \textbf{70.61} & \textbf{70.61} & \textbf{57.88} \\
      & COT+FICL      & 27.75 & 70.73 & 70.73 & 56.40 \\
    \midrule
    \multirow{4}{*}{Llama-3.1-8B-Instruct}
      & GIP           & 9.74  & 70.44 & 70.44 & 50.20 \\
      & COT           & 10.19 & 70.47 & 70.47 & 50.37 \\
      & FICL          & 27.16 & 69.57 & 69.57 & 55.43 \\
      & COT+FICL      & 26.00 & 69.58 & 69.58 & 55.06 \\
    \midrule
    \multirow{4}{*}{Llama-3.2-3B-Instruct}
      & GIP           & 7.13  & 70.56 & 70.56 & 49.41 \\
      & COT           & 6.14  & 71.32 & 71.32 & 49.60 \\
      & FICL          & 16.07 & 70.21 & 70.21 & 52.16 \\
      & COT+FICL      & 16.98 & 70.70 & 70.70 & 52.79 \\
    \midrule
    \multirow{4}{*}{Llama-3.3-70B-Instruct}
      & GIP           & 7.79  & 71.28 & 71.28 & 50.12 \\
      & COT           & 8.75  & 71.34 & 71.34 & 50.48 \\
      & FICL          & 31.25 & 70.94 & 70.94 & 57.71 \\
      & COT+FICL      & 28.97 & 70.95 & 70.95 & 56.95 \\
    \midrule
    \multirow{4}{*}{Mistral-7B-Instruct-v0.3}
      & GIP           & 6.99  & 71.07 & 71.07 & 49.71 \\
      & COT           & 7.26  & 70.87 & 70.87 & 49.67 \\
      & FICL          & 16.96 & 70.99 & 70.99 & 52.98 \\
      & COT+FICL      & 18.58 & 70.74 & 70.74 & 53.35 \\
    \bottomrule
  \end{tabular}
\end{table*}


\section{Extracting Relationship Pairs from Simple Sentences} 
\label{app:t3}
We went back to our coreference annotators and asked them if they could look at a small sample of the dataset and see if the model was successfully able to parse all relationships
\begin{table*}[ht]
  \centering
  \caption{Model Accuracy on Boolean Relationship Extraction from Simple Sentences}
  \label{tab:prompting-eval4}
  \begin{tabular}{l c c c}
    \toprule
    Model Name                       & \# Params (B) & Accuracy   & F$_1$-Score \\
    \midrule
    LLaMA-3-8B                       & 8             & 98.00\%    & 98.00\%     \\
    Mistral-7B                       & 7             & 87.00\%    & 93.55\%     \\
    DeepSeek-Distilled-LLaMA-8B      & 8             & 62.00\%    & 77.02\%     \\
    Qwen-7B                          & 7             & 52.00\%    & 68.42\%     \\
    LLaMA-2-7B                       & 7             & 35.00\%    & 51.85\%     \\
    QwenChat-7B                      & 7             & 23.00\%    & 37.40\%     \\
    DeepSeek-7B                      & 7             & 21.00\%    & 34.71\%     \\
    DeepSeek-Prover-7B               & 7             & 20.00\%    & 33.61\%     \\
    DeepSeek-Distilled-Qwen-7B       & 7             & 11.00\%    & 19.82\%     \\
    \textbf{Mistral-MoE-8$\times$7B} & \textbf{8$\times$7} & \textbf{99.00\%} & \textbf{99.50\%} \\
    DeepSeek-67B                     & 67            & 43.00\%    & 60.14\%     \\
    LLaMA-2-13B                      & 13            & 13.00\%    & 23.01\%     \\
    GPT-NeoX-20B                     & 20            &  0.00\%    &  0.00\%     \\
    LLaMA-2-70B                      & 70            & 41.00\%    & 58.57\%     \\
    \bottomrule
  \end{tabular}
\end{table*}

\begin{table*}[ht]
  \centering
  \scriptsize
  \caption{Performance comparison across models and prompting styles}
  \label{tab:prompt_performance}
  \begin{tabular}{llcccc}
    \toprule
    \textbf{Model} & \textbf{Prompting Style} & \textbf{Macro Average} & \textbf{Exact-Match} & \textbf{RMSE} \\
    \midrule
    \multirow{4}{*}{LLAMA 3 8B} 
      & \textbf{COT+FICL}  & 99.78\% & 98.00\% & 0.1078 \\
      & COT       & 82.86\% & 64.00\% & 0.4265 \\
      & FICL      & 68.89\% & 28.33\% & 0.5376 \\
      & GIP       & 45.81\% & 27.33\% & 0.9319 \\
    \midrule
    \multirow{4}{*}{MISTRAL 8 BY 7}
      & \textbf{COT+FICL}  & 96.14\% & 91.67\% & 0.2323 \\
      & COT       & 92.42\% & 83.00\% & 0.3146 \\
      & FICL      & 90.23\% & 78.00\% & 0.3585 \\
      & GIP      & 82.26\% & 57.67\% & 0.4563 \\
    \midrule
    \multirow{4}{*}{MISTRAL 7 B}
      & \textbf{COT+FICL}  & 96.64\% & 90.33\% & 0.2356 \\
      & COT       & 84.72\% & 64.33\% & 0.4280 \\
      & FICL     & 85.78\% & 67.00\% & 0.4016 \\
      & GIP       & 78.88\% & 53.00\% & 0.5055 \\
    \midrule
    \multirow{4}{*}{LLAMA 3 70B}
      & \textbf{COT+FICL}  & 95.30\% & 81.00\% & 0.3213 \\
      & COT       & 86.39\% & 68.33\% & 0.4064 \\
      & FICL      & 71.98\% & 45.33\% & 0.6064 \\
      & GIP       & 62.01\% & 35.00\% & 0.7829 \\
    \bottomrule
  \end{tabular}
\end{table*}

\label{app:ablation}
\begin{table*}[t]            
  \centering                 
  \caption{Triple-extraction performance on the evaluation set}
  \begin{tabular}{lrrrr}
    \toprule
    Configuration & Triples & Precision & Recall & F1 Score \\
    \midrule
    Human Standard                             & 398 & 100.00\% & 100.00\% & 100.00\% \\
    Full Model (Ours)                          & 422 &  92.00\% &  92.90\% &  92.40\% \\
    \quad-- Remove Coref Resolution            & 376 &  80.60\% &  74.60\% &  77.50\% \\
    \quad-- Remove Sentence Decomposition      & 208 &  74.60\% &  46.20\% &  57.20\% \\
    \quad-- Remove Coref + Sentence Decomposition & 220 & 76.80\% & 42.70\% & 54.80\% \\
    DeepSeek R1                                & 323 &  93.50\% &  78.60\% &  85.40\% \\
    ChatGPT 4o                                 & 215 &  98.10\% &  52.80\% &  68.50\% \\
    NotebookLM                                 &  67 & 100.00\% &  16.83\% &  28.82\% \\
    ChatGPT 4.5                                & 238 &  99.58\% &  59.55\% &  74.53\% \\
    \bottomrule
  \end{tabular}
  \label{tab:triple_results}
\end{table*}

\twocolumn
\FloatBarrier
\begin{table}[!ht]
  \centering
  \scriptsize            
  \setlength{\tabcolsep}{4pt}  
  \caption{Co-reference Group AB and Group CD, where each group’s “link” set is the intersection of its two annotators.}
  \label{tab:group_ab_cd_agreement}
  \resizebox{\columnwidth}{!}{%
    \begin{tabular}{l c}
      \toprule
      \textbf{Statistic}                        & \textbf{Value} \\
      \midrule
      \multicolumn{2}{l}{\it Group definitions:  
         A and B = 2,041; 
         C and D = 1,929} \\
      \midrule
      Number of “link” assignments (Group AB)  & 2,041         \\
      Number of “link” assignments (Group CD)  & 1,929         \\
      Intersection                             & 1,847         \\
      Union                                    & 2,123         \\
      \midrule
      Observed agreement \(P_o\)               & $\approx$ 0.87 \\
      Expected agreement estimated \(P_e\)     & 0.50          \\
      Cohen’s \(\kappa\)                       & 0.74          \\
      \bottomrule
    \end{tabular}
  }
\end{table}
\FloatBarrier

\section{Prompting Strategy}
\subsection{Prompt Templates for Co-reference Resolution}      
\label{app:corefprompt}
\begin{tcolorbox}[
  title=\textbf{COT+FICL Co-reference Resolution Prompt},
  colback=gray!5,
  colframe=black!75,
  fonttitle=\bfseries,
  breakable
]


You are a coreference resolution agent. Below is a biomedical abstract presented as tokenized text with indices. Your task is to identify and annotate coreference expressions within the text. For each co-referent expression:

\begin{itemize}
\item Record the surface form under ``Expression''.
\item Use the provided token indices as ``StartToken'' and ``EndToken'' (they are the same for single-token expressions).
\item Map each expression to its antecedent using ``RefersTo'' --- either a noun phrase or named entity from the text.
\item Only include pronouns or repeated noun phrases referring back to a prior concept or entity.
\end{itemize}

Use this format:
\begin{lstlisting}[basicstyle=\small\ttfamily]
{
"Expression": "string",
"StartToken": int,
"EndToken": int,
"RefersTo": "string"
}
\end{lstlisting}

\textbf{Example:}

Given this tokenized abstract:

\begin{lstlisting}[basicstyle=\small\ttfamily]
("BACKGROUND:", 0), ("There", 1), 
("are", 2), ("few", 3), ("cases", 4), 
("of", 5), ("pulmonary", 6), 
("granulomatous", 7), ("changes", 8), 
("secondary", 9), ("to", 10), 
("primary", 11), ("biliary", 12), 
("cirrhosis", 13), ("(PBC).", 14), 
("No", 15), ("case", 16), ("of", 17), 
("granulomatous", 18), ("lung", 19), 
("disease", 20), ("secondary", 21), 
("to", 22), ("PBC", 23), 
("misdiagnosed", 24), ("as", 25), 
("lung", 26), ("cancer", 27), 
("had", 28), ("been", 29), 
("reported.", 30), ("CASE", 31), 
("SUMMARY:", 32), ("A", 33), 
("middle-aged", 34), ("woman", 35), 
("presented", 36), ("with", 37), 
("lung", 38), ("nodules", 39), 
("and", 40), ("was", 41), 
("misdiagnosed", 42), ("with", 43), 
("lung", 44), ("cancer", 45), 
("by", 46), ("positron", 47), 
("emission", 48), 
("tomography/computed", 49), 
("tomography.", 50), ("She", 51), 
("underwent", 52), ("left", 53), 
("lobectomy,", 54), ("and", 55), 
("the", 56), ("pathology", 57), 
("of", 58), ("the", 59), 
("nodules", 60), ("showed", 61), 
("granulomatous", 62), 
("inflammation,", 63), ("which", 64), 
("was", 65), ("then", 66), 
("treated", 67), ("with", 68), 
("antibiotics.", 69), 
("However,", 70), ("a", 71), 
("new", 72), ("nodule", 73), 
("appeared.", 74), ("Further", 75), 
("investigation", 76), ("with", 77), 
("lung", 78), ("biopsy", 79), 
("and", 80), ("liver", 81), 
("serology", 82), ("led", 83), 
("to", 84), ("the", 85), 
("diagnosis", 86), ("of", 87), 
("PBC,", 88), ("and", 89), 
("chest", 90), ("computed", 91), 
("tomography", 92), 
("indicated", 93), 
("significant", 94), 
("reduction", 95), ("in", 96), 
("the", 97), ("pulmonary", 98), 
("nodule", 99), ("by", 100), 
("treatment", 101), ("with", 102), 
("methylprednisolone", 103), 
("and", 104), 
("ursodeoxycholic", 105), 
("acid.", 106), 
("CONCLUSION:", 107), 
("Diagnosis", 108), ("of", 109), 
("pulmonary", 110), ("nodules", 111),
("requires", 112), 
("integrating", 113), 
("various", 114), ("clinical", 115), 
("data", 116), ("to", 117), 
("avoid", 118), ("unnecessary", 119), 
("pulmonary", 120), 
("lobectomy.", 121)
[
{
"Expression": "PBC",
"StartToken": 14,
"EndToken": 14,
"RefersTo": "Primary 
biliary cirrhosis"
},
{
"Expression": "PBC",
"StartToken": 23,
"EndToken": 23,
"RefersTo": "Primary biliary 
cirrhosis"
},
{
"Expression": "She",
"StartToken": 51,
"EndToken": 51,
"RefersTo": "A middle-aged 
woman"
},
{
"Expression": "PBC",
"StartToken": 88,
"EndToken": 88,
"RefersTo": "Primary biliary 
cirrhosis"
}
]
\end{lstlisting}


Now process this tokenized abstract:

\texttt{\{tokenized\_text\}}

\end{tcolorbox}

\begin{tcolorbox}[
  title=\textbf{COT Co-reference Resolution Prompt},
  colback=gray!6,
  colframe=black!75,
  fonttitle=\bfseries,
  breakable
]


You are a coreference resolution agent. Below is a biomedical abstract presented as tokenized text with indices. Your task is to identify and annotate coreference expressions within the text. For each co-referent expression:

\begin{itemize}
\item Record the surface form under ``Expression''.
\item Use the provided token indices as ``StartToken'' and ``EndToken'' (they are the same for single-token expressions).
\item Map each expression to its antecedent using ``RefersTo'' --- either a noun phrase or named entity from the text.
\item Only include pronouns or repeated noun phrases referring back to a prior concept or entity.
\end{itemize}

Use this format:
\begin{lstlisting}[basicstyle=\small\ttfamily]
{
"Expression": "string",
"StartToken": int,
"EndToken": int,
"RefersTo": "string"
}
\end{lstlisting}


Now process this tokenized abstract:

\texttt{\{tokenized\_text\}}






\end{tcolorbox}

\begin{tcolorbox}[
  title=\textbf{FICL Co-reference Resolution Prompt},
  colback=gray!5,
  colframe=black!75,
  fonttitle=\bfseries,
  breakable
]


You are a coreference resolution agent. Below is a biomedical abstract presented as tokenized text with indices. Your task is to identify and annotate coreference expressions within the text. 

Use this format:
\begin{lstlisting}[basicstyle=\small\ttfamily]

{
"Expression": "string",
"StartToken": int,
"EndToken": int,
"RefersTo": "string"
}
\end{lstlisting}

\textbf{Example:}

Given this tokenized abstract:
\begin{lstlisting}[basicstyle=\small\ttfamily]
("BACKGROUND:", 0), ("There", 1), 
("are", 2), ("few", 3), ("cases", 4), 
("of", 5), ("pulmonary", 6), 
("granulomatous", 7), ("changes", 8), 
("secondary", 9), ("to", 10), 
("primary", 11), ("biliary", 12), 
("cirrhosis", 13), ("(PBC).", 14), 
("No", 15), ("case", 16), ("of", 17), 
("granulomatous", 18), ("lung", 19), 
("disease", 20), ("secondary", 21), 
("to", 22), ("PBC", 23), 
("misdiagnosed", 24), ("as", 25), 
("lung", 26), ("cancer", 27), 
("had", 28), ("been", 29), 
("reported.", 30), ("CASE", 31), 
("SUMMARY:", 32), ("A", 33), 
("middle-aged", 34), ("woman", 35), 
("presented", 36), ("with", 37), 
("lung", 38), ("nodules", 39), 
("and", 40), ("was", 41), 
("misdiagnosed", 42), ("with", 43), 
("lung", 44), ("cancer", 45), 
("by", 46), ("positron", 47), 
("emission", 48), 
("tomography/computed", 49), 
("tomography.", 50), ("She", 51), 
("underwent", 52), ("left", 53), 
("lobectomy,", 54), ("and", 55), 
("the", 56), ("pathology", 57), 
("of", 58), ("the", 59), 
("nodules", 60), ("showed", 61), 
("granulomatous", 62), 
("inflammation,", 63), ("which", 64), 
("was", 65), ("then", 66), 
("treated", 67), ("with", 68), 
("antibiotics.", 69), 
("However,", 70), ("a", 71), 
("new", 72), ("nodule", 73), 
("appeared.", 74), ("Further", 75), 
("investigation", 76), ("with", 77), 
("lung", 78), ("biopsy", 79), 
("and", 80), ("liver", 81), 
("serology", 82), ("led", 83), 
("to", 84), ("the", 85), 
("diagnosis", 86), ("of", 87), 
("PBC,", 88), ("and", 89), 
("chest", 90), ("computed", 91), 
("tomography", 92), 
("indicated", 93), 
("significant", 94), 
("reduction", 95), ("in", 96), 
("the", 97), ("pulmonary", 98), 
("nodule", 99), ("by", 100), 
("treatment", 101), ("with", 102), 
("methylprednisolone", 103), 
("and", 104), 
("ursodeoxycholic", 105), 
("acid.", 106), 
("CONCLUSION:", 107), 
("Diagnosis", 108), ("of", 109), 
("pulmonary", 110), ("nodules", 111),
("requires", 112), 
("integrating", 113), 
("various", 114), ("clinical", 115), 
("data", 116), ("to", 117), 
("avoid", 118), ("unnecessary", 119), 
("pulmonary", 120), 
("lobectomy.", 121)

[
{
"Expression": "PBC",
"StartToken": 14,
"EndToken": 14,
"RefersTo": "Primary biliary 
cirrhosis"
},
{
"Expression": "PBC",
"StartToken": 23,
"EndToken": 23,
"RefersTo": "Primary biliary 
cirrhosis"
},
{
"Expression": "She",
"StartToken": 51,
"EndToken": 51,
"RefersTo": "A middle-aged 
woman"
},
{
"Expression": "PBC",
"StartToken": 88,
"EndToken": 88,
"RefersTo": "Primary biliary 
cirrhosis"
}
]
\end{lstlisting}


Now process this tokenized abstract:

\texttt{\{tokenized\_text\}}

\end{tcolorbox}

\begin{tcolorbox}[
  title=\textbf{GIP Co-reference Resolution Prompt},
  colback=gray!5,
  colframe=black!75,
  fonttitle=\bfseries,
  breakable
]


You are a coreference resolution agent. Below is a biomedical abstract presented as tokenized text with indices. Your task is to identify and annotate coreference expressions within the text. 
Use this format:

Use this format:
\begin{lstlisting}[basicstyle=\small\ttfamily]
{
"Expression": "string",
"StartToken": int,
"EndToken": int,
"RefersTo": "string"
}
\end{lstlisting}


Now process this tokenized abstract:

\texttt{\{tokenized\_text\}}



\end{tcolorbox}

\subsection{Annotators Details}
\label{app:corefannotation}
Annotator A is working on a medical degree
Annotator B is a linguistic
Annotator C is a biologist
Annotator D is a linguistic

Annotators A and B are grouped to work together and produce a perfect work and Annotators C and D are also grouped in a similar pattern.

Annotator E, F and G all were tested before given the code and had a score of 93\% in a specialized test different from the normal tests before being accepted for review. Annotators H and I, are post-graduate students. 
All annotators are from Nigeria. 

They were all recruited by a recruitment expert Sophia Anuyah. They all spoke English and submitted their resumes and were invited for an online interview. 

The annotators were paid in their local currency weekly, at an average of 40,000 NGN a week over the period of 5 weeks. The project was self-funded by the authors.

\subsection{Creating the Sentence Structure Data Set}
\label{app:sentence}
Three initial annotators - Annotator E, B and F selected 7,500 sentences spanning five syntactic categories (compound-complex, compound, complex, simple, incomplete).  Then two senior experts - G and H selected from a cohort of 12 people who took a classification test and scored above 93\% were selected and then adjudicated these and reached consensus on 7,269 sentences out of 7,500 sentences making a 96.92\% agreement rate. The authors chose not to resolve the 231 disagreements due to the current size of the dataset. The final dataset comprises of 2,118 (29.1\%) compound-complex, 1,191 (16.4\%) compound, 865 (11.9\%) complex, 1,585 (21.8\%) simple, and 1,510 (20.8\%) incomplete sentences. From this set, we drew a balanced training sample of 2,000 sentences (400 per class) and reserved the remaining 5,269 sentences for testing. The dataset is available on github.

\section{Prompt for Sentence Conversion}
\subsection{Converting Complex Sentences to Simple Sentences}
\label{app:cx}
Given the sentence: 
\begin{quote}
“A prospective cohort study was conducted in Leeds, UK, based on routinely collected data from a service that allowed patients with symptoms of lung cancer to request CXR” \cite{bradley2021estimating}. 
\end{quote}

\noindent
The process in this conversion is to identify the singular independent clause and the other dependent clauses
\begin{itemize}
    \item Independent Clause: “A prospective cohort study was conducted in Leeds, UK."
    \item Dependent Clauses and Modifiers: (a) “that allowed patients with symptoms of lung cancer to request CXR.” (b) “based on routinely collected data from a service”
\end{itemize}

\noindent
In this example, there was one dependent clause, and one modifier which in our context still depends on the subject for RE, hence, they are looped together in our prompt. Once the LLM can correctly identify the independent clause, the next stage would be parse each relationship separately meaning we have three simple sentences i.e. (1) the independent clause (2) the subject of the IC and the DC and (3) the subject of the IC and the modifier. In this case:


\begin{equation}
\small
\begin{split}
S 
&= \underbrace{\text{(A prospective cohort study was conducted in Leeds, UK)}}_{S_{\mathrm{main}}} \\
&\quad\cup\;
  \underbrace{\text{(that allowed patients … to request CXR)}}_{DC}\,.
\end{split}
\end{equation}

We define an extraction operator \( \mathcal{E}(\cdot) \) that maps a complex sentence to a set of simple (independent) sentences:


\begin{multline}
\mathcal{E}(S_{\text{complex}}) 
= \bigl\{\,S_{\mathrm{main}},\;\text{R}(DC_1),\;
  \text{R}(DC_2),\,\ldots\bigr\}\,.
\end{multline}
where R $\rightarrow$ Rewrite

\begin{equation}
S = S_{\mathrm{main}} \cup DC
\end{equation}
 we get:

$$
\mathcal{E}(S) =
\left\{
\text{S1},
\text{S2},\;
\text{S3}
\right\}.
$$

\begin{quote}
$S1 \rightarrow $ A prospective cohort study was conducted in Leeds, UK.
$S2 \rightarrow $ The study was based on routinely collected data from a service.
$S3 \rightarrow $ The service allowed patients with symptoms of lung cancer to request CXR
\end{quote}

\subsection{Converting Compound Sentences to Simple Sentences}
\label{app:cd}
Given the sentence:
\begin{quote}
“Lung cancer stands prominently among the foremost contributors to human mortality, distinguished by its elevated fatality rate and the second-highest incidence rate among malignancies, and the metastatic dissemination of lung cancer stands as a primary determinant of its elevated mortality and recurrence rates.”
\end{quote}

\noindent
Our goal is to break down this compound sentence into simpler, stand-alone statements.

\begin{itemize}
    \item Independent Clause 1 (IC1): “Lung cancer stands prominently among the foremost contributors to human mortality.”
    \item Independent Clause 2 (IC2): “The metastatic dissemination of lung cancer stands as a primary determinant of its elevated mortality and recurrence rates.”
    \item Dependent Modifier (DM): “distinguished by its elevated fatality rate and the second-highest incidence rate among malignancies” 
\end{itemize}

\noindent
Here, IC1 and IC2 are connected by a coordinating conjunction (i.e., “and”), which is typical in compound sentences. The phrase “distinguished by its elevated fatality rate ... among malignancies” modifies “Lung cancer” (from IC1).

\medskip
\noindent
To convert this compound sentence into simple sentences, we isolate each clause, ensuring each stands alone:

\[
S = \left( IC1 \,\cup\, DM \,\cup\, IC2 \right)
\]

\noindent
We define an extraction operator \( \mathcal{E}(\cdot) \) that maps a compound sentence \( S_{\text{compound}} \) to a set of simple (independent) sentences:

\[
\mathcal{E}(S_{\text{compound}}) 
= 
\left\{
S_{1},\;
S_{2},\;
S_{3}
\right\}.
\]

\noindent
Applying it to our sentence:

\[
S = \left( IC1 \,\cup\, DM \,\cup\, IC2 \right)
\]

\noindent
we obtain three simple sentences:

\begin{quote}
\noindent
$S1 \rightarrow$ Lung cancer stands prominently among the foremost contributors to human mortality.\\
$S2 \rightarrow$ It is distinguished by its elevated fatality rate and the second-highest incidence rate among malignancies.\\
$S3 \rightarrow$ The metastatic dissemination of lung cancer stands as a primary determinant of its elevated mortality and recurrence rates.
\end{quote}

\subsection{Converting Compound-Complex Sentences to Simple Sentences}
\label{app:cc}
Given the sentence:
\begin{quote}
“Although lung cancer is the leading cause of US cancer-related deaths, lung cancer screening with a low radiation dose chest computed tomography scan is now standard of care for a high-risk eligible population, and clinicians and surgeons must evaluate the trade-offs of benefits and harms, including the identification of many benign lung nodules, overdiagnosis, and complications.”
\end{quote}

\begin{itemize}
    \item Independent Clause 1 (IC1): “Lung cancer screening with a low radiation dose chest computed tomography scan is now standard of care for a high-risk eligible population”
    \item Independent Clause 2 (IC2): “Clinicians and surgeons must evaluate the trade-offs of benefits and harms, ”
    \item Dependent Clause (DC): “Although lung cancer is the leading cause of US cancer-related deaths” 
    \item Modifiers: "including the identification of many benign lung nodules, overdiagnosis, and complications"
\end{itemize}

\noindent
We see that DC modifies or sets a contrasting context for IC1, and IC1 is coordinated with IC2 via “and.”
\medskip
\noindent
To convert this into simple sentences, each clause (or key part of a clause) should form its own standalone statement:

\[
S = \left( DC \,\cup\, IC1 \,\cup\, IC2 \right)
\]

\noindent
Using our extraction operator \( \mathcal{E}(\cdot) \):

\[
\mathcal{E}(S_{\text{compound-complex}}) 
= 
\left\{
S_{1},\;
S_{2},\;
S_{3},\; ..., S_{n}
\right\},
\]
\noindent
we obtain:
\begin{quote}
\noindent
$S1 \rightarrow$ Lung cancer is the leading cause of US cancer-related deaths.\\
$S2 \rightarrow$ Lung cancer screening with a low-dose chest computed tomography scan is now standard of care for a high-risk eligible population.\\
$S3 \rightarrow$ Lung cancer screening is recommended for a high-risk, eligible population. \\
$S4 \rightarrow$ Clinicians and surgeons must evaluate the trade-offs of benefits and harms, \\
$S5 \rightarrow$ Evaluated trade-offs of benefits and harms include the identification of many benign lung nodules. \\
$S6 \rightarrow$ Evaluated trade-offs of benefits and harms include the risk of over-diagnosis.	\\
$S7 \rightarrow$ Evaluated trade-offs of benefits and harms include complications from lung-cancer screening 
\end{quote}


\subsection{Prompts}
\label{app:cot+ficl}
\begin{tcolorbox}[
  title=\textbf{COT+FICL Complex Sentence Conversion},
  colback=gray!5,
  colframe=black!75,
  fonttitle=\bfseries,
  breakable
]

Below is a step-by-step process.  For each example, think step by step, then output only the simplified sentences in the form:
\[
  \text{S1} \;\rightarrow\;\dots\quad
  \text{S2} \;\rightarrow\;\dots\quad
  \dots
\]
one per line, and nothing else.

\bigskip
\noindent\textbf{Example 1:}\\
\textbf{Input:}
\begin{quote}
“A prospective cohort study was conducted in Leeds, UK, based on routinely collected data from a service that allowed patients with symptoms of lung cancer to request CXR.”
\end{quote}
\textbf{Chain-of-Thought:}
\begin{enumerate}
  \item Identify the independent clause:  
    “A prospective cohort study was conducted in Leeds, UK.”
  \item Identify dependent clauses/modifiers:  
    \begin{itemize}
      \item Modifier A: “based on routinely collected data from a service”  
      \item Dependent clause B: “that allowed patients with symptoms of lung cancer to request CXR”
    \end{itemize}
  \item Rewrite each as a standalone simple sentence:
\end{enumerate}
\textbf{Output:}
\begin{itemize}
  \item S1 \(\rightarrow\) A prospective cohort study was conducted in Leeds, UK.
  \item S2 \(\rightarrow\) The study was based on routinely collected data from a service.
  \item S3 \(\rightarrow\) The service allowed patients with symptoms of lung cancer to request CXR.
\end{itemize}

\bigskip
\noindent\textbf{Example 2:}\\
\textbf{Input:}
\begin{quote}
“After the cells were treated with the drug, which had been synthesized in our lab, we measured the change in fluorescence using a spectrophotometer.”
\end{quote}
\textbf{Chain-of-Thought:}
\begin{enumerate}
  \item Independent clause:  
    “We measured the change in fluorescence using a spectrophotometer.”
  \item Dependent clauses/modifiers:  
    \begin{itemize}
      \item Dependent clause A: “After the cells were treated with the drug”  
      \item Modifier B: “which had been synthesized in our lab”
    \end{itemize}
  \item Rewrite each as standalone simple sentences:
\end{enumerate}
\textbf{Output:}
\begin{itemize}
  \item S1 \(\rightarrow\) We measured the change in fluorescence using a spectrophotometer.
  \item S2 \(\rightarrow\) The cells were treated with the drug.
  \item S3 \(\rightarrow\) The drug had been synthesized in our lab.
\end{itemize}

\bigskip
\noindent\textbf{Now apply the same process to this new sentence:}

\textbf{Input:}  
\verb|"{{sentence}}"|

\medskip
\noindent\textbf{***OUTPUT ONLY the simplified sentences, one per line in the form S1 → …, S2 → …, etc., and nothing else.***}

Now process this abstract:

```ABSTRACT GIVEN HERE'''

\end{tcolorbox}

\begin{tcolorbox}[
  title=\textbf{COT+FICL Compound Sentence Conversion},
  colback=gray!5,
  colframe=black!75,
  fonttitle=\bfseries,
  breakable
]
Below is a process to convert a compound sentence into simple sentences.
For each example, think step by step, then output only the simplified sentences in the form:
\[
  \text{S1} \rightarrow \dots
  \quad
  \text{S2} \rightarrow \dots
  \quad
  \dots
\]
one per line, and nothing else.

\medskip
\noindent\textbf{Example 1:}\\
\textbf{Input:}
\begin{quote}
“Lung cancer stands prominently among the foremost contributors to human mortality, distinguished by its elevated fatality rate and the second-highest incidence rate among malignancies, and the metastatic dissemination of lung cancer stands as a primary determinant of its elevated mortality and recurrence rates.”
\end{quote}
\textbf{Chain-of-Thought:}
\begin{enumerate}
  \item Identify the independent clauses:
    \begin{itemize}
      \item IC1: “Lung cancer stands prominently among the foremost contributors to human mortality.”
      \item IC2: “The metastatic dissemination of lung cancer stands as a primary determinant of its elevated mortality and recurrence rates.”
    \end{itemize}
  \item Identify modifiers:
    \begin{itemize}
      \item Modifier: “distinguished by its elevated fatality rate and the second-highest incidence rate among malignancies” (modifies IC1)
    \end{itemize}
  \item Rewrite all parts as simple, standalone sentences.
\end{enumerate}
\textbf{Output:}
\begin{itemize}
  \item S1 \(\rightarrow\) Lung cancer stands prominently among the foremost contributors to human mortality.
  \item S2 \(\rightarrow\) It is distinguished by its elevated fatality rate and the second-highest incidence rate among malignancies.
  \item S3 \(\rightarrow\) The metastatic dissemination of lung cancer stands as a primary determinant of its elevated mortality and recurrence rates.
\end{itemize}

\medskip
\noindent\textbf{Example 2:}\\
\textbf{Input:}
\begin{quote}
“Climate change accelerates the melting of polar ice, and rising sea levels threaten coastal communities around the world.”
\end{quote}
\textbf{Chain-of-Thought:}
\begin{enumerate}
  \item Identify the independent clauses:
    \begin{itemize}
      \item IC1: “Climate change accelerates the melting of polar ice.”
      \item IC2: “Rising sea levels threaten coastal communities around the world.”
    \end{itemize}
  \item No dependent clauses or modifiers.
  \item Rewrite each as a standalone simple sentence.
\end{enumerate}
\textbf{Output:}
\begin{itemize}
  \item S1 \(\rightarrow\) Climate change accelerates the melting of polar ice.
  \item S2 \(\rightarrow\) Rising sea levels threaten coastal communities around the world.
\end{itemize}

\medskip
\noindent\textbf{Now apply the same process to this new sentence:}\\
\textbf{Input:}  
\verb|"{{sentence}}"|  

\medskip
\noindent\textbf{OUTPUT ONLY the simplified sentences, one per line in the form S1 → …, S2 → …, etc., and nothing else.}
\end{tcolorbox}


\begin{tcolorbox}[
  title=\textbf{COT+FICL Compound-Complex Sentence Conversion},
  colback=gray!5,
  colframe=black!75,
  fonttitle=\bfseries,
  breakable
]
Below is a process to split a compound-complex sentence into standalone simple sentences.  Think step by step, then apply.

\medskip
\noindent\textbf{Example 1:}\\
\textbf{Input:}
\begin{quote}
“Although lung cancer is the leading cause of US cancer-related deaths, lung cancer screening with a low radiation dose chest computed tomography scan is now standard of care for a high-risk eligible population, and clinicians and surgeons must evaluate the trade-offs of benefits and harms, including the identification of many benign lung nodules, overdiagnosis, and complications.”
\end{quote}
\textbf{Chain-of-Thought:}
\begin{enumerate}
  \item Dependent Clause (DC): “Although lung cancer is the leading cause of US cancer-related deaths”
  \item Independent Clause 1 (IC1): “Lung cancer screening with a low-dose chest computed tomography scan is now standard of care for a high-risk eligible population”
  \item Independent Clause 2 (IC2): “Clinicians and surgeons must evaluate the trade-offs of benefits and harms”
  \item Modifier list: “including the identification of many benign lung nodules, overdiagnosis, and complications”
  \item Rewrite into standalone simple sentences.
\end{enumerate}
\textbf{Output:}
\begin{itemize}
  \item S1 $\rightarrow$ Lung cancer is the leading cause of US cancer-related deaths.
  \item S2 $\rightarrow$ Lung cancer screening with a low-dose chest computed tomography scan is now standard of care for a high-risk eligible population.
  \item S3 $\rightarrow$ Lung cancer screening is recommended for a high-risk, eligible population.
  \item S4 $\rightarrow$ Clinicians and surgeons must evaluate the trade-offs of benefits and harms.
  \item S5 $\rightarrow$ Evaluated trade-offs include the identification of many benign lung nodules.
  \item S6 $\rightarrow$ Evaluated trade-offs include the risk of overdiagnosis.
  \item S7 $\rightarrow$ Evaluated trade-offs include complications from lung-cancer screening.
\end{itemize}

\bigskip
\noindent\textbf{Example 2:}\\
\textbf{Input:}
\begin{quote}
“Although warmed by the sun, the fields remained dry, and farmers worried about the drought.”
\end{quote}
\textbf{Chain-of-Thought:}
\begin{enumerate}
  \item Dependent Clause (DC): “Although warmed by the sun”
  \item Independent Clause 1 (IC1): “The fields remained dry”
  \item Independent Clause 2 (IC2): “Farmers worried about the drought”
  \item Rewrite into standalone simple sentences.
\end{enumerate}
\textbf{Output:}
\begin{itemize}
  \item S1 $\rightarrow$ The sun warmed the fields.
  \item S2 $\rightarrow$ The fields remained dry.
  \item S3 $\rightarrow$ Farmers worried about the drought.
\end{itemize}

\bigskip
\noindent\textbf{Now apply to this new sentence:}\\
\textbf{Input:}  
\verb|"{{sentence}}"|  

\medskip
\noindent\textbf{OUTPUT ONLY the simplified sentences, one per line in the form S1 $\rightarrow$ …, S2 $\rightarrow$ …, etc., and nothing else.}
\end{tcolorbox}

\begin{tcolorbox}[
  title=\textbf{COT + FICL Relationship Extraction for Knowledge Graph},
  colback=gray!5,
  colframe=black!75,
  fonttitle=\bfseries,
  breakable
]
You are a knowledge graph relationship extraction agent. Your task is to extract structured relationships from simple sentences to create knowledge graph triples. Each triple should contain two entities and the relationship between them.

\begin{itemize}
\item Analyze the sentence structure and identify key components.
\item Extract all meaningful entities (nouns, noun phrases, proper nouns, concepts).
\item Identify relationships between entities based on verbs, prepositions, and semantic meaning.
\item Form triples (Entity 1 → Relationship → Entity 2) as structured relationships.
\item Validate that each triple captures meaningful semantic information.
\end{itemize}

\textbf{Examples:}

\begin{quote}"Regulating miR-497-5p provides a potential targeted therapy for lung cancer treatment."
\end{quote}

\begin{lstlisting}[basicstyle=\small\ttfamily]
[{
"Entity 1": "regulating miR-497-5p", 
"Entity 2": "lung cancer targeted 
treatment",
"Relationship": "provides"
}]
\end{lstlisting}

\begin{quote}"The activation of caspase signal pathway was the reason for stronger apoptosis."
\end{quote}

\begin{lstlisting}[basicstyle=\small\ttfamily]
[{
"Entity 1": "activation of caspase 
signal pathway",
"Entity 2": "stronger apoptosis",
"Relationship": "was the reason for"
}]
\end{lstlisting}

\begin{quote}"With clinical significance features selection, over-sampling methods achieved the highest AUC results."
\end{quote}

\begin{lstlisting}[basicstyle=\small\ttfamily]
[{
"Entity 1": "clinical significance 
features selection",
"Entity 2": "over-sampling methods",
"Relationship": "With"},
{"Entity 1": "over-sampling methods",
"Entity 2": "highest AUC results",
"Relationship": "achieved"}]
\end{lstlisting}

Now extract knowledge graph relationships from this sentence: 
\texttt{\{sentence\}}

\end{tcolorbox}

All other prompt types are in our code base.

\newpage
\begin{figure*}[ht]
  \centering
  \includegraphics[width=1\linewidth]{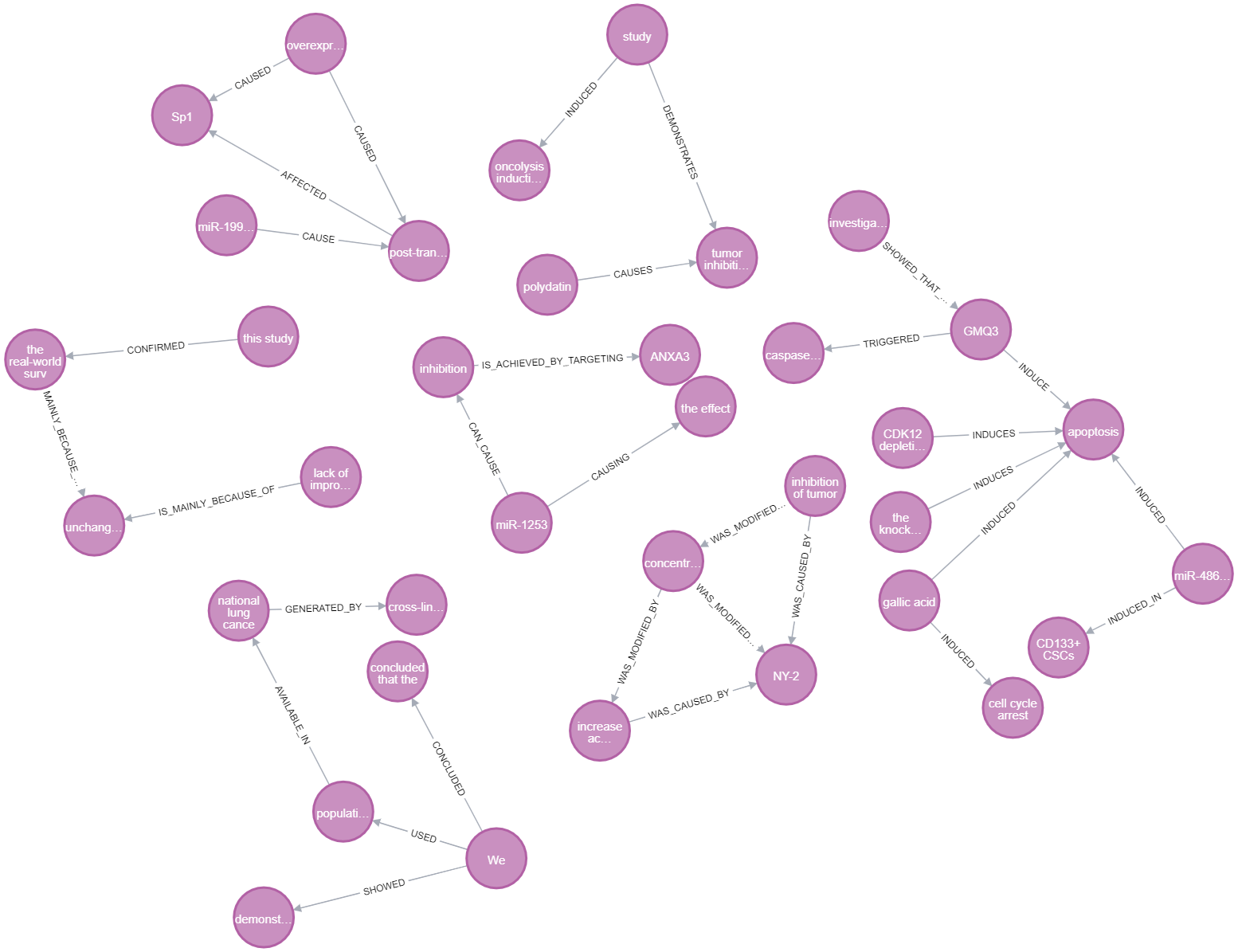}
  \captionsetup{justification=centering} 
  \caption{Subsection of the Knowledge Graph.}
  \label{fig:graph}
\end{figure*}



\end{document}